\documentclass[5p,times]{elsarticle}
\usepackage{ecrc}
\usepackage{hyperref}       % hyperlinks
\usepackage{url}    
\usepackage{afterpage}
\usepackage{booktabs}       % professional-quality tables
\usepackage{siunitx}
\usepackage{subcaption}
\usepackage{tabularx}
\usepackage{siunitx}
\usepackage{amsmath}
\usepackage[table]{xcolor}
\usepackage{enumitem}
\usepackage{caption}
\usepackage{times}
\usepackage{graphicx}
\usepackage{scalerel}
\usepackage{fancyhdr}
\usepackage{stfloats}
\usepackage{lastpage}
\usepackage{bbm}
\usepackage{csquotes}
\usepackage{svg}
\usepackage{float}
\usepackage{placeins}
\usepackage[figuresright]{rotating}
\usepackage{tabularx}
\usepackage{makecell}
\usepackage{threeparttable}   % put this in the preamble
\usepackage{multirow}
\usepackage{microtype}
\volume{00}

\makeatletter

\def\ps@pprintTitle{%
  \def\@oddhead{%
    \setbox1=\hbox{\elslogo}%
    \setbox2=\hbox{\sdlogo}%
    \setbox3=\hbox{\jnllogo}%
    \vspace*{2pc}
    \parbox[t]{\wd1}{\elslogo}%
    \hfil
    \parbox[t]{19pc}{\centering%
      \raisebox{23pt}{\sdlogo}\\[-12pt]
    }
    \hfil
    \raisebox{23pt}{\parbox[c]{\wd3}{\jnllogo\\[6pt]}}%
  }
  \let\@oddfoot\@empty
  \let\@evenhead\@oddhead
  \let\@evenfoot\@oddfoot
}
\makeatother

\firstpage{1}

\runauth{J. Henrich et al.}

\usepackage{amssymb}
\usepackage{lineno}
\newcommand{\ch}[1]{\textcolor{black}{#1}}

\newcommand{\revchanged}[1]{\textcolor{black}{#1}}
\newcommand{\fig}[1]{Fig.~\ref{#1}}
\newcommand{\secref}[1]{Sec.~\ref{#1}}
\newcommand{\tab}[1]{Tab.~\ref{#1}}

\newcommand{\pigemoji}{%
  \raisebox{-0.15ex}{\includegraphics[height=1.8ex]{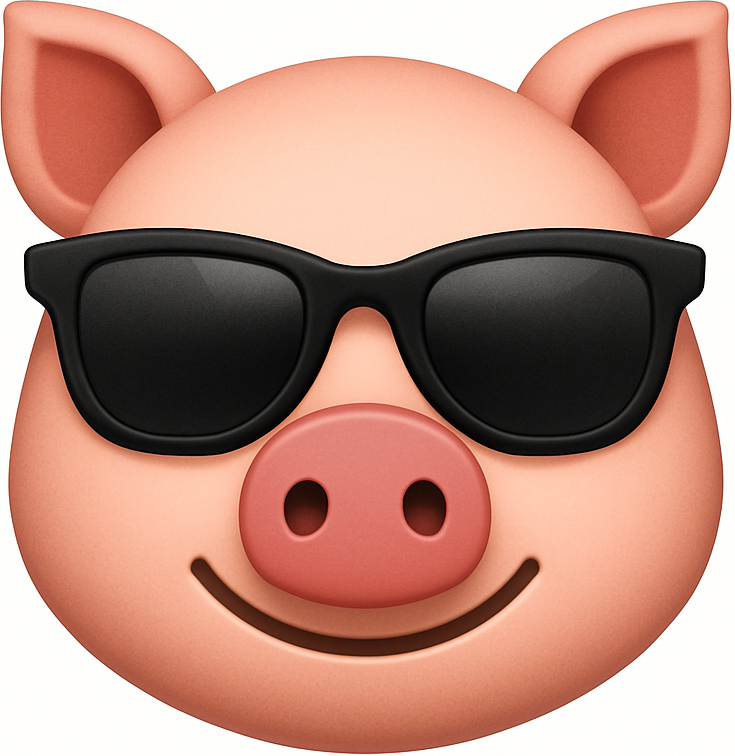}}%
}

\newcommand{\TP}{\text{TP}_{\alpha}}
\newcommand{\FP}{\text{FP}_{\alpha}}
\newcommand{\FN}{\text{FN}_{\alpha}}
\newcommand{\deta}{\text{DetA}_{\alpha}}
\newcommand{\assa}{\text{AssA}_{\alpha}}
\newcommand{\hota}{\text{HOTA}_{\alpha}}

\newcommand{\TPA}{\text{TPA}_{\alpha}}
\newcommand{\FPA}{\text{FPA}_{\alpha}}
\newcommand{\FNA}{\text{FNA}_{\alpha}}
\newcommand{\TPAC}{\text{TPA}_{\alpha}\text{(}c\text{)}}
\newcommand{\FPAC}{\text{FPA}_{\alpha}\text{(}c\text{)}}
\newcommand{\FNAC}{\text{FNA}_{\alpha}\text{(}c\text{)}}

\begin{document}
\label{lastpage}
\begin{frontmatter}

\dochead{}
    
\title{Benchmarking pig detection and tracking under diverse and challenging conditions \pigemoji}

% % Authors, for the paper (add full first names)
\author[thomas,jonathan]{Jonathan Henrich\corref{corr}}
\author[imke]{Christian Post}
\author[goettingen]{Maximilian Zilke}
\author[goettingen]{Parth Shiroya}
\author[goettingen]{Emma Chanut}
\author[kiel]{Amir Mollazadeh Yamchi}
\author[ramin,jonathan]{Ramin Yahyapour}
\author[thomas,jonathan]{Thomas Kneib}
\author[imke]{Imke Traulsen}

\cortext[corr]{Correspondence: jonathan.henrich@uni-goettingen.de}
\address[thomas]{Chair of Statistics and Econometrics, Faculty of Economics, University of Göttingen, Germany}
\address[jonathan]{Campus Institute Data Science}
\address[imke]{Institute of Animal Breeding and Husbandry, Kiel University, Germany}
\address[goettingen]{work performed during employment at University of Göttingen, Germany}
\address[kiel]{work performed during employment at Kiel University, Germany}
\address[ramin]{Gesellschaft für wissenschaftliche Datenverarbeitung mbH Göttingen, Germany}

\begin{abstract}
To ensure animal welfare and effective management in pig farming, monitoring individual behavior is a crucial prerequisite. While monitoring tasks have traditionally been carried out manually, advances in machine learning have made it possible to collect individualized information in an increasingly automated way. Central to these methods is the localization of animals across space (object detection) and time (multi-object tracking). Despite extensive research of these two tasks in pig farming, a systematic benchmarking study has not yet been conducted. In this work, we address this gap by curating two datasets: PigDetect for object detection and PigTrack for multi-object tracking. The datasets are based on diverse image and video material from realistic barn conditions, and include challenging scenarios such as occlusions or bad visibility. For object detection, we show that challenging training images improve detection performance beyond what is achievable with randomly sampled images alone. Comparing different approaches, we found that state-of-the-art models offer substantial improvements in detection quality over real-time alternatives. For multi-object tracking, we observed that SORT-based methods achieve superior detection performance compared to end-to-end trainable models. However, end-to-end models show better association performance, suggesting they could become strong alternatives in the future. We also investigate characteristic failure cases of end-to-end models, providing guidance for future improvements. The detection and tracking models trained on our datasets perform well in unseen pens, suggesting good generalization capabilities. This highlights the importance of high-quality training data. The datasets and research code are made publicly available to facilitate reproducibility, re-use and further development.

\end{abstract}


\end{frontmatter}
\pagestyle{plain}
% \linenumbers

\section{Introduction}
Monitoring individual behavior in pig farming is essential to ensure a high level of animal welfare and the efficient functioning of work processes. Traditionally, the acquisition of behavioral information has been time-consuming and laborious, as it relied on human observation and documentation. In the last decade, this has started to change due to the emergence of powerful machine learning methods that allow to automate this task. Automatic methods for individual identification \cite{t_psota_long-term_2020, wutke2025multistage} and individualized action understanding \cite{tong2022videomae, wu2021towards} have the potential to serve as digital assistance tools for farmers and researchers to monitor the health status \cite{alameer_automated_2020, bhujel_deep-learning-based_2021}, to detect potentially harmful behaviors like tail biting\cite{liu_computer_2020} or mounting\cite{yang2021pig, li_mounting_2019}, or to observe normal behavior with the goal of detecting changes as early \revchanged{indicators of health and welfare problems} \cite{bergamini_extracting_2021, riekert_automatically_2020, hesse2023computer}. At the core of such methods lies the localization of individual animals across space and time, commonly referred to as \textit{object detection} and \textit{multi-object tracking} in the technical literature. Since these tasks often are modular components in more sophisticated behavioral analysis pipelines \cite{vogg2025computer}, establishing strong, robust and readily available methods for pig detection and tracking is a crucial prerequisite for further developments. While detection and tracking methods based on fine-grained instance representations such as segmentation masks \cite{he2017mask, wang_end--end_2021}, or keypoints exist \cite{toshev2014deeppose, mathis2018deeplabcut, t_psota_long-term_2020}, most of the technical literature is based on axis-aligned bounding boxes. Similarly, state-of-the-art computer vision methods for individualized action understanding primarily rely on bounding boxes \cite{gu2018ava, gritsenko_end--end_2023-1, ryali2023hiera, wu_memvit_2022}. This predominance can be attributed to the ease of annotation and the availability of increasingly powerful backbones \cite{ryali2023hiera, dosovitskiy_image_2021} that allow tracking- and action-related information to be extracted directly from images or videos without the need for complex intermediate representations. Although rotated bounding boxes \cite{liu2017learning, liu2023s} can more accurately capture the extent of objects, they are incompatible with the vast majority of multi-object tracking and action understanding methods. As a result, it is difficult to benefit from future advances in these areas when using this representation. For these reasons, our work focuses on detection and tracking methods based on axis-aligned bounding boxes.

Many studies in recent years have addressed the problem of pig detection \cite{alameer_automated_2020, riekert_automatically_2020, riekert2021model, hao_novel_2022, ji_automatic_2022, guo_enhanced_2023, mattina_efficient_2023, melfsen2023describing, pu2024tr} and tracking \cite{cowton_automated_2019, zhang2019automatic, alameer_automated_2020, gan_automated_2021, shirke2021tracking, bergamini_extracting_2021, wang_towards_2022, guo_enhanced_2023, parmiggiani2023don, melfsen2023describing, wei2023detection, yang2024long, jaoukaew2024robust, tu2024tracking, yu2025fto, liu2025sdgtrack}, often adapting standard methods to better suit the conditions of pig farming, and achieving promising results. Unfortunately, only a few authors made their detection \cite{riekert_automatically_2020, riekert2021model, alameer_automated_2020, melfsen2023describing} and tracking datasets \cite{bergamini_extracting_2021, melfsen2023describing, jaoukaew2024robust, yu2025fto} publicly available. These datasets are limited in terms diversity, often including only a single barn or pen environment, and have not become well established as benchmark datasets, as they are rarely used for comparisons across different works. Other works state that research data is available upon request. However, we did not receive responses to the inquiries we made, suggesting that access to such data may be limited in practice. Similarly, user-friendly code bases for pig detection and tracking are scarce, as research code is often not made publicly available within the livestock research community. Recent works on pig detection and tracking are beginning to acknowledge this gap by making both their code and datasets publicly available \cite{melfsen2023describing, yu2025fto}.

The lack of publicly accessible code bases and datasets is, in our opinion, problematic for two reasons: (1) Current methods cannot be compared to determine the state of the art. Many studies on pig detection \cite{hao_novel_2022, ji_automatic_2022, guo_enhanced_2023, mattina_efficient_2023, pu2024tr} and tracking \cite{cowton_automated_2019, zhang2019automatic, alameer_automated_2020, gan_automated_2021, wang_towards_2022, guo_enhanced_2023, parmiggiani2023don, wei2023detection, yang2024long, tu2024tracking} report evaluation metrics on undisclosed datasets, making direct comparisons between them impossible. In the broader field of computer vision, methods are usually benchmarked on standardized, publicly available datasets \cite[e.\,g.][]{deng2009imagenet, lin2014microsoft, gu2018ava}, which facilitates reproducibility and encourages fair performance comparisons. (2) The lack of accessible resources also hampers the development and the availability of robust pig detection and tracking methods which are urgently needed for downstream analysis tasks. Researchers that need to localize animals as part of their research often start by re-inventing the wheel. They annotate large amounts of data and adjust generic code bases for detection and tracking to train and hyperparameter-tune their own models without benefiting from the work of researchers that already tackled this task. In contrast, open data and open source are common practice in the broader field of computer vision, such that methods can be easily fine-tuned, extended or used as modular components in custom analysis pipelines.
The widespread availability of open-source code bases and high-quality labeled data has been a major driver of the advancements in computer vision in the last decade, and would greatly benefit precision livestock farming research as well.

In this work, we aim to address this gap by curating two datasets: \textit{PigDetect} for object detection and \textit{PigTrack} for multi-object tracking. Both datasets contain samples from diverse barn environments in pig farming that were annotated with bounding boxes. They include realistic conditions, such as pigs occluding each other, bad lighting or smudged camera lenses. For PigDetect, we specifically included challenging images in the dataset by identifying and correcting cases where a trained pig detection model commits errors.

We benchmarked the performance of several recent general-purpose detection and tracking models on our datasets. In addition to models that are commonly used in pig farming, such as YOLO variants for object detection or SORT-based methods for multi-object tracking, we also evaluated models that have not yet been employed in this context. Specifically, we included Co-DINO \cite{zong2023detrs}, a detection model that achieves state-of-the-art results on the COCO dataset \cite{lin2014microsoft}, and two recent end-to-end trainable models for multi-object tracking \cite{zhang_motrv2_2023, gao2025multiple} that achieve state-of-the-art results on the DanceTrack dataset \cite{sun2022dancetrack}. Our comprehensive benchmarking study enables a systematic comparison between methods and highlights strengths and weaknesses of different model classes. For the end-to-end models we also conducted a detailed error analysis, providing guidance for future improvements.

\revchanged{We further evaluated both the detection and tracking models on a third-party dataset from a previously unseen pen environment, where they demonstrated strong performance. In particular, the tracking models substantially outperformed the best results previously reported by \citet{yu2025fto}.} This highlights the importance of carefully selected, high-quality training data for the development of robust and generalizable models. PigDetect (\url{https://doi.org/10.25625/I6UYE9}) and PigTrack (\url{https://doi.org/10.25625/P7VQTP}), along with the trained model weights and the source code for training, evaluation and inference (\url{https://github.com/jonaden94/PigBench}), are made publicly available to facilitate reproducibility, re-use and further development. In summary, our study has three main contributions:

\begin{itemize}
\item A diverse and challenging benchmark dataset for both pig detection and tracking.
\item A comparison of the performance of several state-of-the-art general-purpose detection and tracking models on our datasets.
\item Carefully documented source code for training, evaluation and inference of the models employed in this work.
\end{itemize}

\section{PigDetect dataset}\label{sec:pig}

\subsection{Annotation}\label{sec:pig_ann}
The images in our PigDetect dataset come from previously unpublished video footage recorded in the projects with grant numbers 28RZ372060 (Inno Pig), 2817902315 (FriSch), EIP-Agri-Ni-2019-LPTT-03 (Old Breed New House), and 2822ZR015 (TiPP), as well as image data published in prior works \cite{psota_multi-pig_2019, bergamini_extracting_2021, alameer_automated_2022} that had not yet been annotated with bounding boxes. \revchanged{By including third-party sources, we aimed to further increase the diversity of the dataset.} To obtain images for annotation, 2363 images were randomly sampled from the available image (N~=~975) and video sources (N~=~1388). For the video sources, frames were extracted as sparsely as possible to ensure diversity. Specifically, we randomly sampled distinct one-hour videos and subsequently extracted one frame from each of these videos. The sampled frames were then annotated with axis-aligned bounding boxes using the labelme tool \cite{wada_labelme_2023} and form the foundation of our benchmark dataset. 

Some challenging detection scenarios cannot be adequately captured with random sampling because they occur too infrequently. To identify images as “challenging”, we trained a pig detector (YOLOX, \cite{ge_yolox_2021}) with the randomly sampled images and visually inspected the predicted bounding boxes for a large number of images from all sources using the interactive labeling tool pylabel \cite{heaton_pylabel_2023}. If the predicted bounding boxes were considered to contain severe errors (no formal criterion), the annotation was corrected. \revchanged{The most frequent error cases were missed detections (e.\,g. due to occlusion or poor visibility), multiple bounding boxes for a single pig (e.\,g. for a pig split into two separate visible parts through occlusion), and highly inaccurate bounding boxes (e.\,g. two pigs erroneously detected as one in low-contrast scenarios). By identifying and correcting erroneous cases,} another 568 images were added to the dataset from the image (N~=~252) and video sources (N~=~316), resulting in a total of 2931 images. Of these images, 1704 were sampled from the previously unpublished video footage (Inno Pig, FriSch, Old Breed New House, TiPP), and 1227 from the other sources \cite{psota_multi-pig_2019, bergamini_extracting_2021, alameer_automated_2022}. 

The annotation was performed by one person and independently reviewed by another, with corrections applied where needed. Two of the authors were involved in this procedure (J.~H. and C.~P.). It should be noted that some images could not be unambiguously annotated (e.\,g., cases of heavy piling) and were therefore not included in the dataset, which highlights the practical limitations of pig detection.

\subsection{Characteristics}\label{sec:pig_cha}
\begin{figure*}[t]        % ! = override constraints, h = right here
  \centering
  \includegraphics[width=\textwidth]{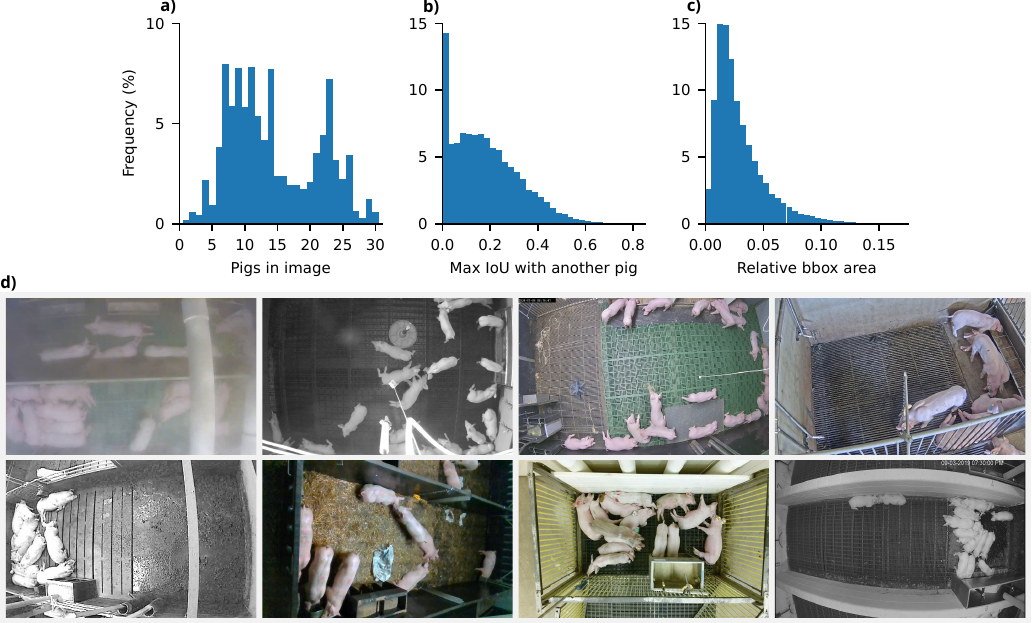}
  \caption{Distributional properties of ground truth annotations and example images of PigDetect. (a) displays the distribution of the number of pigs per image across all images. The bin width is set to one. (b) shows the distribution of a pig’s maximum intersection over union (IoU) with another pig in the same image across all pigs in all images. The bin width is set to 0.025. (c) shows the distribution of a pig's bounding box size relative to image size across all pigs in all images. The bin width is set to 0.005. For subfigures b and c, large values that occur too rarely to be visible in the histograms were excluded. (d) depicts example images from PigDetect that include challenging conditions such as occlusions and bad visibility. The top row shows images from Inno Pig, Old Breed New House, TiPP, and  FriSch (left to right). The bottom row shows images from \citet{alameer_automated_2022}, \citet{bergamini_extracting_2021}, and two from \citet{psota_multi-pig_2019} (left to right).}
  \label{fig:pig-detect-stats}
\end{figure*}

\begin{table}[t]
    \centering
    \caption{Dataset characteristics of PigDetect compared to publicly available pig object detection datasets. Previous works did not distinguish between random and challenging images, so the field is left blank. \revchanged{The number of images that are both challenging and captured during night is 97 (train), 34 (val) and 27 (test).}}
    \label{tab:pigdetect}
    \footnotesize
    \setlength{\tabcolsep}{2.5pt}
    \begin{tabular}{llllll}
        \toprule
        Dataset & Images & Challenging & Night & BBoxes & Pens \\
        \midrule
        \citet{riekert_automatically_2020, riekert2021model}   & 582 & - & 156 & \num[group-digits=integer]{13020} & 18 \\
        \citet{alameer_automated_2020}   & \num[group-digits=integer]{113379} & - & 0 & \num[group-digits=integer]{735094} & 1 \\
        \citet{melfsen2023describing}   & 614 & - & 0 & 6754 & 1 \\
        \midrule
        PigDetect (our)   & 2931 & 568 & 687 & \num[group-digits=integer]{42177} & 31 \\
        train & 2431 & 414 & 581 & \num[group-digits=integer]{33197} & 30 \\
        val   &  250 &  34 &  66 &  3544 & 30 \\
        test  &  250 & 120 &  40 &  5436 & 1 \\
        \bottomrule
    \end{tabular}
\end{table}

%Overview
\tab{tab:pigdetect} provides dataset characteristics of PigDetect compared to other publicly available pig detection datasets. Previously published works employed only random sampling for dataset curation without specifically selecting challenging images, which we show to be suboptimal (see \secref{res:chall}). Furthermore, PigDetect contains more images and annotated bounding boxes than most previously published datasets \cite{riekert_automatically_2020, riekert2021model, melfsen2023describing}. \citet{alameer_automated_2020} published an action understanding dataset that contains images from conventional pig farming annotated with bounding boxes that are associated with an action label. While this dataset is more extensive than PigDetect, it contains only daytime images from a single pen. Compared to all other datasets, PigDetect includes a larger number of visually diverse pens \ch{from conventional indoor barn environments} (\fig{fig:pig-detect-stats}d) that were captured during day and night. A comprehensive depiction of the visual diversity of the pen settings of previously published datasets (\fig{fig:app_det_third}) and PigDetect (\fig{fig:app_det_ours}) can be found in \ref{app:sec_det}. For a more detailed breakdown of image characteristics for the different sources included in PigDetect, we refer to \tab{tab:source_stats}.

%Quantitative
Image resolutions in our dataset vary greatly, with the smallest and largest images having a width and height of $630 \times 355$ pixels and $3840 \times 2160$ pixels respectively. This is neither a problem for object detection nor for tracking since training and inference pipelines usually include automatic resizing and padding. For night images, most sources used infrared imaging. In total, there are 687 night images in the dataset (23~\%). The data coming from one barn of \citet{alameer_automated_2022} consists of images where only an infrared camera was used during the day. These 153 images appear as grayscale and were thus counted as night images.

%Qualitative
The majority of pigs found in the images from the video sources are crossbreed pigs, (German Large White \texttimes\ German Landrace) \texttimes\ (Piétrain or Duroc). Images from the project \enquote{Old Breed New House} contain Swabian Hallish, and Bentheim crossbreed pigs. The animals in the images from \citet{alameer_automated_2022} are either Duroc \texttimes\ (Landrace \texttimes\ Large White) or Danbred Duroc \texttimes\ (Danbred Yorkshire \texttimes\ Danbred Landrace). \citet{psota_multi-pig_2019} and \citet{bergamini_extracting_2021} did not disclose the breed of their pigs, but they appear to be mostly typical crossbreeds, with a few exceptions in \citet{psota_multi-pig_2019} that include pigs with spotted or belted colors, as well as a few pigs that appear to be Poland China pigs. Throughout the dataset, pigs were captured at different stages of development from rearing to fattening. 
% Pigs were captured in from 30 kg body weight until slaughter; too Precise since we do not have this info for other sources

% Environments
% eigene Bilder

Images taken from \citet{psota_multi-pig_2019} are the most diverse in terms of backgrounds, containing 17 different pens. The images in \citet{alameer_automated_2022} are from multiple pens in two different barn locations, and in \citet{bergamini_extracting_2021} only a single pen background is present. The data from Inno Pig, FriSch, Old Breed New House and TiPP (see \secref{sec:pig_ann}) contains 11 distinct environments that differ in floor type and pen equipment.

The pens were captured from above, often with multiple cameras at different angles. Most of the images were taken in a practical barn environment. This resulted in various conditions and image characteristics that pose a challenge to object detectors. These include low contrast due to insufficient lighting or overexposure, smudges or dust on the camera lens, motion blur, and partial occlusion of pigs through other pigs or barn interior. One reason for severe occlusions is piling of pigs, which is also included in the dataset to the extent that it could still be annotated. Some images do not fully capture the pen they are monitoring \revchanged{or contain parts of other pens}, which results in pigs that are only partially visible at the edges of the image. \revchanged{If these pigs could still be clearly identified, they were annotated as well.}

% whose maximum IoU with the bounding box of another pig is substantially larger than zero
The distribution of the number of pigs in an image has local maxima around 10 and 23 (\fig{fig:pig-detect-stats}a)\revchanged{. The minimum number of pigs per video is one, and the maximum is 32}. \revchanged{To quantify the extent of overlap between pigs, we computed for each bounding box the maximum intersection-over-union (IoU) with any other bounding box (\fig{fig:pig-detect-stats}b).} The resulting distribution shows that the bounding boxes of a large number of pigs have little to no overlap with those of other pigs, reflecting the fact that pigs are often lying or standing around alone to sleep, rest, eat or drink. \revchanged{However, it is also evident that most pigs in the dataset have bounding boxes that overlap with those of others.} Bounding box sizes relative to the image size vary widely from very small bounding boxes that comprise less than 1~\% of the image up to larger bounding boxes that comprise 10~\% of the image (\fig{fig:pig-detect-stats}c). Example images from diverse barn environments that also showcase challenging detection scenarios are depicted in \fig{fig:pig-detect-stats}d.

\subsection{Dataset splits}\label{sec:det_splits}

% train test split
Images from one barn (N~=~250) were selected as the test set and show an environment that is not present in the training (N~=~2431) and validation data (N~=~250). This independence of our test set was chosen to challenge the robustness of models, as well as simulate practical conditions where models have to perform in previously unseen environments. Apart from randomly sampled images (N~=~130, 13 night images), the test set contains a substantial portion of challenging images (N~=~120, 27 night images). We chose to include a large proportion of challenging images to enable a better differentiation between the performance of different methods. \tab{tab:pigdetect} includes the number of images, challenging images, night images, and annotated bounding boxes for each dataset split. A depiction of the pen environment of the test set can be found in \fig{fig:app_det_ours} (pen setting H6).

\section{PigTrack dataset}

\subsection{Annotation}
The videos in our PigTrack dataset originate from the same video footage that has been used to obtain images for PigDetect (see \secref{sec:pig_ann}). \revchanged{At the time of annotation, we were not aware of publicly available video datasets with a license that permits republication of modified versions, which is why we did not enrich the diversity of our dataset with third-party videos.} To obtain videos for annotation, we first sighted the footage to specifically identify segments that featured challenging tracking conditions (no formal criterion). \revchanged{This included scenarios with rapid, non-linear movements, poor visibility, and occlusions. In case of temporary occlusions, it was ensured that all pigs could still be unambiguously identified throughout the video. For example, this is the case if only one pig is occluded at a time, if a pig reappears in the same location, or if it follows a clear trajectory during the occlusion. To avoid further ambiguities, it was also ensured that no pig leaves the scene and re-enters afterwards. Pigs leaving without returning or new pigs entering throughout the video was considered acceptable.}

The selected video segments were input to a basic tracking method to generate preliminary annotations. Specifically, we used YOLOX \cite{ge_yolox_2021} to generate frame-wise bounding boxes and linked them using a simple motion-based matching algorithm (C-BIoU, \cite{yang_hard_2023}). The resulting preliminary annotations were loaded into CVAT \cite{cvat}, an image and video annotation tool, so that only the errors had to be manually corrected. This protocol greatly improved annotation efficiency since correctly predicted bounding boxes and \mbox{(sub-)}tracks could be left unchanged, allowing annotators to focus on refining only the challenging parts. \ch{CVAT furthermore allows hiding all tracks except one, such that the videos could be annotated one pig at a time, which prevents sensory overload.} Tracks were annotated by one person and independently reviewed by a second person, with corrections applied if necessary. Five of the authors were involved in this procedure (J.~H., C.~P., P.~S., E.~C. and A.~M.~Y.). In total, 80 videos were densely annotated (every frame), spanning 41.06~min of material. As with PigDetect, some particularly complex cases could not be unambiguously annotated and were therefore not included in the dataset.

\subsection{Characteristics}\label{sec:track_char}
\begin{figure*}[t]        % ! = override constraints, h = right here
  \centering
  \includegraphics[width=\textwidth]{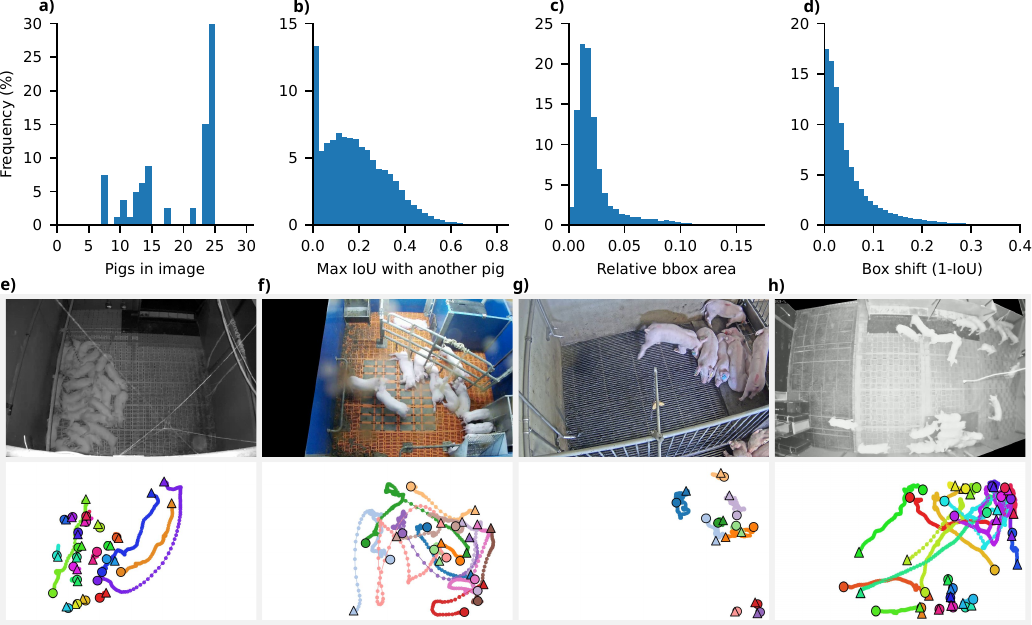}
  \caption{Distributional properties of ground truth annotations and examples of PigTrack. (a) displays the distribution of the number of pigs per video across all videos. The bin width is set to one. (b) shows the distribution of a pig’s maximum IoU with another pig in the same video frame across all pigs in all video frames. The bin width is set to 0.025. (c) shows the distribution of a pig's bounding box size relative to video frame size across all pigs in all video frames. The bin width is set to 0.005. (d) displays the distribution of bounding box shifts of the same individual from one frame to the next across all pigs in all video frames. The bin width is set to 0.01. For subfigures b, c, and d, large values that occur too rarely to be visible in the histograms were excluded. (e-h) show example sequences from PigTrack. The top row displays the first frame of the corresponding video, while the bottom row illustrates the trajectories of individual pigs over time, with each pig represented by a distinct color. Each trajectory starts with a circle and ends with a triangle.}
  \label{fig:pig-track-stats}
\end{figure*}

\begin{table}[t]
    \centering
    % \todo{Actually, the Liu et al. dataset is also only available via password and I did not receive an answer yet.. So I could not calculate number of IDs and bboxes and also could not evaluate on the dataset.}
    \caption{Dataset characteristics of PigTrack compared to publicly available pig tracking datasets. \textit{Type} denotes the annotation type, i.e. whether every frame was annotated with bounding boxes (dense) or only a certain number of key-frames (sparse).}
    \label{tab:pigtrack}
    \scriptsize
    \setlength{\tabcolsep}{1pt}
    \begin{tabular}{lrrrrrrrr}
        \toprule
        Dataset &T (min) & Videos & Night & Frames & \ch{Tracks} & BBoxes & Pens & Type \\
        \midrule
        \citet{bergamini_extracting_2021} &60.00 &  12 & 0 & \num[group-digits=integer]{21600} & 96 & \num[group-digits=integer]{12646} & 1 & sparse \\
        \citet{melfsen2023describing} &14.00 &  14 & 3 & \num[group-digits=integer]{25200} & 154 & 9240 & 1 & sparse \\
        \citet{jaoukaew2024robust} & 9.33&  4 & 0 & \num[group-digits=integer]{19339} & 41 & 554 & 1 & sparse \\
        \citet{yu2025fto} & 2.50 &  1 & 0 & 1260 & 23 & \num[group-digits=integer]{28814} & 1 & dense \\
        % \citet{liu2025sdgtrack} & 10.00 & 10 & 2 & \num[group-digits=integer]{18000} & NA & NA & 9 & dense \\
        \midrule
        PigTrack (our) & 41.06 &  80 & 26 & \num[group-digits=integer]{26487} & 1574 & \num[group-digits=integer]{517961} & 9 & dense\\
        train & 15.12 & 35 & 13 & 9959  & 698  & \num[group-digits=integer]{195745} & 9 & dense\\
        val  & 12.06 & 20 & 3 & 7589  & 383  & \num[group-digits=integer]{141289} & 7 & dense\\
        test  & 13.88 &25 & 10 & 8939  & 493  & \num[group-digits=integer]{180927} & 9 & dense\\
        \bottomrule
    \end{tabular}
\end{table}

%Overview
\tab{tab:pigtrack} summarizes the characteristics of PigTrack in comparison to existing pig tracking datasets. PigTrack contains substantially more videos, night recordings, \ch{tracks}, and annotated bounding boxes than any previously published dataset. While some earlier datasets have longer durations \cite{bergamini_extracting_2021} or a comparable number of frames \cite{bergamini_extracting_2021, melfsen2023describing, jaoukaew2024robust}, they are only sparsely annotated, resulting in far fewer bounding box labels. Specifically, \citet{bergamini_extracting_2021} annotated frames irregularly based on when behaviors change, \citet{melfsen2023describing} annotated only one frame per second, and \citet{jaoukaew2024robust} only every 500th frame. Standard benchmarks for human multi-object tracking are usually densely labeled \cite{milan2016mot16, sun2022dancetrack}. While sparsely labeled bounding boxes can be interpolated for training or evaluation purposes, the quality of tracks will inevitably decrease compared to dense annotations. Moreover, existing pig tracking datasets \cite{bergamini_extracting_2021, melfsen2023describing, jaoukaew2024robust, yu2025fto} are limited to a single pen environment. A comprehensive depiction of the visual diversity of the pen settings of previously published datasets (\fig{fig:app_track_third}) and PigTrack (\fig{fig:app_track_ours}) can be found in \ref{app:sec_track}. For more detailed characteristics for every video included in PigTrack, we refer to \tab{tab:detailed_pigtrack}.

The width and height of most videos (Inno Pig, FriSch, Old Breed New House) is $1280 \times 800$ , while the videos from one source (TiPP) have a resolution of $1920 \times 1080$ pixels. The majority of videos (N~=~62) has 10 frames per second (FPS), while a smaller number operate at either 5~FPS (N~=~2) or 15~FPS (N~=~16). The dataset features nine distinct pen environments and includes 26 night videos, \revchanged{which were captured using infrared illumination}. All pigs visible in a video were annotated. On rare occasions, this included animals in neighboring pens. When pigs in neighboring pens could not be clearly annotated, the corresponding regions were masked by setting their pixel values to zero. \revchanged{If pigs are temporarily invisible in a video sequence (e.\,g., due to occlusion), only those frames in which they are visible were annotated, following established multi-object tracking annotation protocols \cite{milan2016mot16, sun2022dancetrack}.} The breed of pigs as well as the pen and camera characteristics have already been described in \secref{sec:pig_cha}.

Most videos contain between 23 and 26 pigs, while a smaller number contains around 10 pigs (\fig{fig:pig-track-stats}a). The number of pigs per video ranges from a minimum of seven to a maximum of 26. Regarding the maximum IoU of bounding boxes with other bounding boxes in the same video frame and their size relative to the image (\fig{fig:pig-track-stats}b and c), the distributions are similar to the ones reported for PigDetect (see \fig{fig:pig-detect-stats}b and c), which is expected since most of the data was captured in the same setting. To quantify movement, we computed the IoU between a pig’s bounding boxes on adjacent frames and subtracted the result from one. A value of zero indicates no change in the bounding box across frames, while a value of one means the bounding boxes no longer overlap at all. When computing this box shift for every pig in every video frame, we found that most values lie between zero and 0.3, with larger box shifts being rare (\fig{fig:pig-track-stats}d). The trajectories of individual pigs in four example videos from PigTrack are shown in \fig{fig:pig-track-stats}e-h. The videos feature challenging tracking scenarios such as pigs walking over other pigs (\fig{fig:pig-track-stats}e), fast and non-linear movements (\fig{fig:pig-track-stats}f), pigs in close proximity and partially on top of each other (\fig{fig:pig-track-stats}g), and bad visibility paired with occlusions (\fig{fig:pig-track-stats}h).

\subsection{Dataset splits}\label{sec:pigtrack_splits}

Of the 80 existing videos, 35 were assigned to the training set, 20 to the validation set, and 25 to the test set. Unlike PigDetect, the videos in the PigTrack test set were recorded in the same pens as the training and validation videos. To enable a good differentiation between methods, the test set was manually selected to include videos featuring particularly challenging scenarios (no formal criterion) and a large number of night videos (N~=~10). Furthermore, it was verified that the test videos do not contain scenarios that are near-identical to those present in the training and validation sets, such as pigs lying at the same locations in the same pattern. However, there is of course a general similarity of videos from the same pens since pigs tend to spend time in similar places.
Compared to the training and test sets, the validation set contains the least challenging videos. The number of videos, night videos, frames, pigs and bounding boxes as well as the total duration of video material for each dataset split is shown in \tab{tab:pigtrack}.

\section{Benchmarking object detection}

\subsection{Evaluated object detection models} \label{sec:eval_od_methods}

\revchanged{
We benchmarked test performance using the three real-time object detection models YOLOX \cite{ge_yolox_2021}, YOLOv8 \cite{Jocher_Ultralytics_YOLO_2023} and RT-DETR \cite{zhao2024detrs}. Furthermore, we benchmarked the performance of DINO \cite{zhang2022dino} with collaborative hybrid assignments training \cite{zong2023detrs} (Co-DINO), a model focused on detection quality rather than inference speed. These four models were selected for benchmarking because they are widely adopted and have a strong performance. Furthermore, YOLOX, YOLOv8, and Co-DINO are all supported within the same software framework, which we used as the foundation of our code base (see \secref{sec:bench_ob_imp}). Co-DINO achieves state-of-the-art results on the COCO dataset \cite{lin2014microsoft}, but has not yet been employed for pig detection.
}

\subsubsection{\revchanged{YOLOX and YOLO-v8}}

In the original YOLO model proposed by \citet{redmon2016you}, an input image is processed by a convolutional backbone to produce a rich visual feature map of size $7 \times 7 \times 1024$. Intuitively, this can be thought of as corresponding to a $7 \times 7$ grid over the input image, while the feature vector of size 1024 contains all relevant information for objects at the corresponding grid location. The feature vector is subsequently processed by a smaller head network to predict class probabilities and bounding box coordinates relative to the grid position. Compared to more complex proposal-based paradigms, as used by the R-CNN family \cite{girshick2014rich, girshick2015fast, ren2015faster, he2017mask}, the YOLO model is computationally efficient, enabling real-time object detection. 

Successor models like YOLOX \cite{ge_yolox_2021} and YOLOv8 \cite{Jocher_Ultralytics_YOLO_2023} made several improvements over the original model. For example, both models adopt separate head branches for classification and bounding box regression. This helps prevent training conflicts between these semantically distinct tasks by allowing each branch to learn its own features. Notable performance increases have also been achieved by \revchanged{developing more elaborate data} augmentations. For example, multiple images are overlayed (Mixup, \cite{zhang2019bag}) or combined (Mosaic, \cite{bochkovskiy2020yolov4}) to create more challenging prediction scenarios. Other improvements include loss functions that empirically lead to an increased performance \cite{li2020generalized, zheng2020distance} or better label assignment strategies between predicted and ground-truth bounding boxes for more effective training \cite{ge_yolox_2021}.

\subsubsection{\revchanged{Co-DINO and RT-DETR}}\label{sec:codino}

\revchanged{Both Co-DINO and RT-DETR are} based on the seminal DETR \cite{carion2020end} architecture for object detection. In the DETR architecture, a convolutional backbone first extracts features from the input image, which are subsequently refined by a transformer \cite{vaswani2017attention} encoder. Then, a set of learned queries extracts information from these features via cross-attention in a transformer decoder architecture. The transformer decoder also includes self-attention between queries, which is crucial to prevent multiple queries from attending to the same object, thereby avoiding duplicate bounding boxes. Finally, the enriched query features are used for classification and bounding box regression. For supervision, a one-to-one matching needs to be established between queries and ground-truth objects. This is done by choosing the matching that minimizes the loss between the predicted and ground-truth bounding boxes and classes using the Hungarian algorithm \cite{kuhn1955hungarian}. 

The original DETR model suffered from convergence problems and performed worse than fully convolutional object detection approaches. Follow-up works \cite{zhu2020deformable, li2022dn, liu2022dab} proposed several improvements, which have been combined and extended in the DINO model \cite{zhang2022dino}. For example, DINO parameterizes queries as dynamic anchor boxes \cite{liu2022dab}, treating the bounding box regression task as an iterative refinement of these anchors across multiple transformer decoder layers. This spatial informativeness of queries promotes convergence and boosts performance. 

Co-DINO \cite{zong2023detrs} further improves upon the training recipe of DINO. In DINO (and other DETR frameworks), the number of positive queries is limited by the number of objects in an image. This leads to a sparse supervision of the backbone, which is detrimental for feature learning. Co-DINO tackles this by including auxiliary object detection heads such as Faster-RCNN \cite{ren2015faster} for supervision. The small number of positive queries also leads to inefficient cross-attention learning. This is tackled by generating customized positive queries based on bounding boxes obtained from the auxiliary heads. Together, these advancements establish Co-DINO as the state of the art in object detection.

\revchanged{In contrast to Co-DINO and other classical DETR variants, RT-DETR reduces the computational burden of self-attention. Instead of applying self-attention across all backbone features, it restricts it to high-level, semantically rich features, which are then fused with low-level convolutional features. Furthermore, RT-DETR introduces an uncertainty-minimal query selection strategy that yields high-quality object queries by jointly optimizing their classification and localization confidence. Together, these modifications enable real-time inference while preserving competitive detection accuracy.}

\subsubsection{Implementation details} \label{sec:bench_ob_imp}

\begin{table*}[t]
    \centering
    \caption{\revchanged{Model and training specifications for pig detection. \textit{Pretrained Model} indicates the dataset on which the model used for weight initialization was trained. \textit{LR Schedule} denotes how the given learning rate decays over time. For RT-DETR and Co-DINO, the learning rate decays by a factor of 0.1 at the specified step. \textit{LR BB scale} denotes the multiplier applied to the backbone learning rate relative to the base learning rate given by \textit{LR (max/min)}. \textit{Time (h)} denotes the training time.}}
    \label{tab:training_spec}
    \footnotesize
    \begin{tabular}{l l l c c l l l c}
        \toprule
        Model & Backbone & Pretrained Model & Batch Size & Epochs & LR Schedule & LR (max/min) & \revchanged{LR BB Scale} & Time (h) \\
        \midrule
        YOLOX-X   & DarkNet53\cite{redmon2018yolov3}         & COCO\cite{lin2014microsoft}                 & 6  & 115 & Cosine & $\frac{6}{64} \times 10^{-2} \;\big/\; \frac{30}{64} \times 10^{-4}$ & \revchanged{1} & 15.1\\
        YOLOv8-X  & CSP-Darknet\cite{wang2020cspnet}         & COCO                                        & 6  & 110 & Linear & $\frac{6}{64} \times 10^{-2} \;\big/\; \frac{6}{64} \times 10^{-4}$ & \revchanged{1}  & 9.7\\
        \revchanged{RT-DETR-R101} & \revchanged{ResNet\cite{he2016deep}}                 & \revchanged{COCO+Obj365\cite{shao2019objects365}}  & \revchanged{16}  & \revchanged{300} & \revchanged{Step (at 164)} & \revchanged{$1 \times 10^{-4} \;\big/\; 1 \times 10^{-5}$} & \revchanged{0.01} & \revchanged{14.5}\\
        Co-DINO & Swin-L\cite{liu2021swin}                 & COCO+Obj365  & 4  & 100 & Step (at 90) & $\frac{1}{4} \times 10^{-4} \;\big/\; \frac{1}{4} \times 10^{-5}$ & \revchanged{0.1} & 18.5\\
        \bottomrule
    \end{tabular}
\end{table*}

All models were trained and validated according to the dataset splits defined in \secref{sec:det_splits}. The best-performing model in terms of \revchanged{Average Precision} (see \secref{sec:det_eval_metrics}) on the validation set was then used to obtain test performance. YOLOX and YOLOv8 were trained on a single Nvidia A100 GPU using the MMYolo toolbox \cite{mmyolo2022}. \revchanged{RT-DETR was also trained on a single Nvidia A100 GPU using the official GitHub implementation, while} Co-DINO was trained on four A100 GPUs using the MMDetection toolbox \cite{chen2019mmdetection}. All models use an image resolution of $960 \times 960$ pixels. The basic training configurations \revchanged{for Co-DINO and the largest model variants of YOLOX, YOLOv8 and RT-DETR} are listed in \tab{tab:training_spec}. \revchanged{The batch size, number of epochs, and learning rate schedule were determined by experimenting with multiple configurations and selecting the one that achieved the best validation performance. The specified learning rate was obtained by linearly scaling the default value from the original publications according to the chosen batch size.} Other specifications such as learning rate warmup schedules or data augmentations are in line with the original implementations. \revchanged{The training configurations for the smaller model variants of YOLOX, YOLOv8 and RT-DETR partly differ in the learning rate, which was also adopted from the original implementations. Otherwise, they are identical.}

The speed of the trained models was measured on an Nvidia A100 GPU using the inference functionality of the MMDetection toolbox \revchanged{for YOLOX, YOLOv8 and Co-DINO, and using the official PyTorch code for RT-DETR}. \ch{Note that inference speed could be further increased through model export and optimization frameworks such as ONNX Runtime \cite{onnxruntime}. Here we only report the speed of the official research code.} We ran the models on a set of 1000 randomly selected images, repeating the process five times to mitigate variability in performance measurements. The overall median speed per image was computed from these 5000 inferences to obtain a robust assessment. Images were resized in advance to a resolution of $960 \times 960$ pixels to avoid additional overhead. Data loading from disc was not considered in the speed measurements, as it can be avoided in practical applications.

\subsection{Challenging vs. random images: effects on detection} \label{sec:met_cha}
In addition to our benchmarking study, we conducted an experiment with the aim of quantifying differences in performance patterns when using randomly selected vs.\ challenging images during the training of pig detection models. 
The experiment considers the two factors \enquote{number of training images} and \enquote{random vs.\ challenging}. The number of images used for training was varied in nine steps for \enquote{random} (100, 200, 300, 400, 600, 1000, 1400, 1800, 2200) and in four steps for \enquote{challenging} (100, 200, 300, 400). In addition to these variants, the model was also trained with all 2681 training and validation images (random + challenging), resulting in a total of 14 variants. For each variant, five training runs were conducted and the performance was averaged to get a more accurate assessment. In each run, the training images were randomly selected from the pool of all random (N~=~2233) or challenging (N~=~484) images contained in the training and validation set. 

The model used for training is YOLOX \cite{ge_yolox_2021} using a checkpoint trained on COCO as initialization . A constant learning rate of $3/64 \times 10^{-2}$ with a batch size of 3 was used in all 14 variants to ensure comparability. The model was trained with Mosaic and Mixup \revchanged{data augmentations in all epochs}. Otherwise, the training settings were identical to the ones that were used to train the \enquote{YOLOX-L} model in the original YOLOX work \cite{ge_yolox_2021}. Training was terminated when the \revchanged{Average Precision} (see \secref{sec:det_eval_metrics}) on the test set did not increase for 20 consecutive epochs. Performance metrics are reported for the checkpoint with the highest \revchanged{Average Precision} score.

\subsection{Evaluation metrics}\label{sec:det_eval_metrics}

Model performance was evaluated with the commonly used Average Precision metric \cite{everingham_pascal_2010}. 
To construct this metric, a one-to-one matching between predicted and ground-truth bounding boxes is established by minimizing the IoU across all pairs. This is done using the Hungarian algorithm \cite{kuhn1955hungarian}. The algorithm ensures that each predicted bounding box can be matched to at most one ground-truth bounding box and vice versa. The resulting matches are subsequently filtered according to a minimum required IoU of \(\alpha\). This prevents matches of predictions and ground truths that have little or no overlap. Next, all \( N_{\text{pred}} \) predicted bounding boxes in the dataset that is evaluated are ordered by confidence in descending order. Then, precision and recall based on the \( n \) most confident predictions and a specific \( \alpha\) value can be calculated as:

\[
\text{P}_{\alpha}(n) = \frac{\text{M}_{\alpha}(n)}{n}
\]

\[
\text{R}_{\alpha}(n) = \frac{\text{M}_{\alpha}(n)}{N_{\text{gt}}}
\]
where \( \text{M}_{\alpha}(n) \) is the number of matched predictions among the \( n \) most confident predictions and \( N_{\text{gt}} \) is the total number of ground-truth bounding boxes. Once \( \text{P}_{\alpha}(n) \) and \( \text{R}_{\alpha}(n) \) are computed for all \( n\leq N_{\text{pred}} \), the precision-recall curve \( \text{P}_{\alpha}(r) \) (precision as a function of recall) is constructed by plotting \( \text{P}_{\alpha}(n) \) on the vertical axis against \( \text{R}_{\alpha}(n) \) on the horizontal axis, starting with \( n=1 \) (only the most confident prediction is considered) and ending with \(  n=N_{\text{pred}} \) (all predictions are considered). This curve illustrates the trade-off between precision and recall: As the minimum required confidence threshold is lowered, more predictions are included, which typically increases recall at the expense of precision. Finally, the Average Precision metric is computed as the area under the precision-recall curve:

\[
\text{AP}_{\alpha} = \int_0^1 \text{P}_{\alpha}(r) \, dr
\]

Following the literature on object detection, we report two metrics in this work: \(\text{AP}_{0.5}\) and \revchanged{Average Precision} (AP). \(\text{AP}_{0.5}\) is simply \( \text{AP}_{\alpha} \) computed with a minimum required IoU for a match of \( \alpha = 0.5 \). Since the required IoU is relatively low, this metric measures a model's ability to roughly detect objects without taking into account localization accuracy. In contrast, AP is computed by averaging over multiple IoU thresholds:
\[
\text{AP} = \frac{1}{|A|} \sum_{ \alpha \in A} \text{AP}_{\alpha}
\]
where \( A = \{0.50, 0.55, 0.60, \dots, 0.95\} \) is the set of IoU thresholds considered and \(|A|\) denotes the number of elements in the set. Since stricter IoU thresholds are taken into account, this metric measures how precisely objects are localized.

\section{Benchmarking multi-object tracking}

\subsection{Evaluated multi-object tracking models} \label{sec:eval_mot}

We evaluated the tracking performance on the PigTrack test set using several widely adopted methods from the SORT family \cite{bewley2016simple}, ranging from classical approaches like ByteTrack \cite{zhang2022bytetrack} to more recent and stronger variants such as BoT-SORT \cite{aharon2022bot}. These methods rely on bounding box outputs from a pre-trained detector that are linked across frames using motion or appearance-based association rules. In addition to these SORT-based models, we also benchmarked MOTRv2 \cite{zhang_motrv2_2023} and MOTIP \cite{gao2025multiple}, two recent end-to-end trainable trackers, where detection and association are learned in a data-driven way. MOTRv2 and MOTIP achieve state-of-the-art results on the DanceTrack dataset \cite{sun2022dancetrack}, but have not yet been employed for pig tracking.

\subsubsection{SORT family} \label{sec:sort}
In the original SORT model \cite{bewley2016simple}, detections are obtained independently for each frame of a video using Faster R-CNN \cite{ren2015faster}. The detections on the first two frames are matched with each other based on overlap to form object tracks. Then, each object’s motion is calculated as the spatial shift between the corresponding detections in the first two frames. Using these motion estimates, each object’s position in the third frame is predicted by applying the Kalman filter algorithm \cite{kalman1960new}. The actual detections on the third frame are then matched with existing tracks based on their IoU with the predicted object positions. This process is iterated for all frames to obtain tracks for the entire input video. The methods evaluated in this paper improve upon this basic paradigm in several ways. 

\textit{ByteTrack} \cite{zhang2022bytetrack} addresses the fact that automatically discarding low-confidence detections leads to an increased number of false negatives. Therefore, the method uses a two-stage matching strategy. First, detections exceeding a predefined confidence threshold are associated with existing tracks as in the standard SORT paradigm. Then, detections falling below that threshold are associated with tracks that are still unmatched.

\textit{OC-SORT} \cite{cao2023observation} employs mechanisms to counteract noisy predictions that can arise when using the standard Kalman filter. For example, if an object has been occluded for several frames, even small errors in motion estimation can accumulate, leading to largely inaccurate predicted positions. OC-SORT tackles this by updating motion and position estimates based on the new observation once an object is matched again after a period of occlusion. This helps prevent the object from being lost in future frames due to inaccurate position estimates.

\textit{Deep OC-SORT} \cite{maggiolino2023deep} builds upon OC-SORT and improves several key aspects. Most notably, it incorporates a deep association metric, yielding more robust tracking than motion-based association alone. Specifically, a pre-trained Re-ID model is applied on the detected objects to obtain appearance features. These features are then compared with those of existing tracks, and the resulting similarity scores are combined with motion information during the matching process.

\textit{StrongSort} \cite{du2023strongsort} improves DeepSort \cite{wojke2017simple}, which was the first work to incorporate a deep association metric into SORT. Compared to its predecessor, StrongSort uses a more powerful object detector and Re-ID model. Other improvements include the calculation of an exponential moving average of the appearance features of a track. This makes the model less susceptible to noisy detections.

\textit{BoT-SORT} \cite{aharon2022bot} builds upon the ByteTrack framework by incorporating a series of ``bag of tricks'' (hence BoT-SORT) that enhance both motion and appearance modeling. It employs a refined Kalman filter that results in more accurate position estimates. Furthermore, association is strengthened via an adaptive fusion of IoU and appearance similarity, allowing the tracker to dynamically weight motion and appearance cues based on whichever is more informative. BoT-SORT also introduces camera motion compensation to normalize global scene shifts.

\textit{Improved Association} (ImprAssoc \cite{stadler2023improved}) also uses ByteTrack as a foundation and makes several improvements. Instead of using a two-stage matching strategy, all detections are matched in a single step with stricter similarity requirements for low-confidence detections. This allows the most fitting detections to be matched irrespective of their confidence score. Similar to BoT-Sort, ImprAssoc also takes both IoU and appearance similarity into account during matching. Lastly, detections with considerably high overlap are removed, as they likely represent duplicate predictions. This prevents the formation of ghost tracks that do not correspond to real objects.

\subsubsection{End-to-end trainable models} \label{sec:learning_based}

Although the methods described in \secref{sec:sort} include learned components (detector and Re-ID model), they rely on hand-crafted similarity rules for tracking. In contrast, the two approaches presented in this section learn the tracking task in an end-to-end manner, directly optimizing both detection and association through training.

\textit{MOTIP} \cite{gao2025multiple} uses a simple architecture built on top of DETR. Instead of using DETR to only output bounding boxes, it is configured to also output embedding vectors for each object. The embeddings of the first frame are concatenated with learned track ID embeddings and serve as context information for the tracking task. 
% The assignment of objects to IDs is arbitrary. What matters is maintaining consistent associations between each object and its corresponding ID over time.
Object embeddings from the second frame retrieve this context information by interacting with the embeddings from the first frame via cross-attention. The enriched embeddings from the second frame are then used to predict the track ID using a linear layer. For later frames, MOTIP maintains a memory of object embeddings from multiple previous frames, enabling the model to recover identities even after temporary occlusions. Additionally, MOTIP reserves a special ID token that the model should predict whenever a previously unseen object enters the scene. The model can be trained by jointly optimizing the detection objective along with the ID prediction task.

\textit{MOTRv2} \cite{zhang_motrv2_2023} is built on top of the MOTR framework \cite{zeng2022motr}, which also extends DETR for tracking. In MOTR, object embeddings are not only mapped to bounding boxes, but also serve as track queries that contain object-related information. These track queries interact with the image features of the next frame to predict new object embeddings and bounding boxes, enabling frame-by-frame tracking across a video. In addition, separately learned detect queries are used to identify newly appearing objects. The key contribution of MOTRv2 is the incorporation of bounding-box priors from a pre-trained object detector. By allowing track and detect queries to extract information from these priors, the burden of object detection is reduced, enabling MOTRv2 to focus more effectively on association modeling. This leads to a substantial performance increase compared to the MOTR baseline.

\subsubsection{Implementation details}\label{sec:track_implementation}

\begin{table*}[!t]
    \centering
    \caption{\revchanged{Model and training specifications for pig tracking. \textit{Pretrained Model} indicates the dataset on which the model used for weight initialization was trained. The \textit{Step LR schedules} reduce the learning rate by a factor of 0.1 at each step. \textit{LR BB scale} denotes the multiplier applied to the backbone learning rate relative to the base learning rate given by \textit{LR (max/min)}. \textit{Time (h)} denotes the training time. For MOTIP full and MOTRv2, training times are reported for the default setup that only uses the PigTrack training set (see \secref{sec:track_implementation}). For training settings including more data, time scales roughly linearly with the number of frames.}}
    \label{tab:training_spec_tr}
    \footnotesize
    \begin{tabular}{l l l c c l l l c}
        \toprule
        Model & Backbone & Pretrained Model & Batch Size & Epochs & LR Schedule & LR (max/min) & \revchanged{LR BB Scale} & Time (h)\\
        \midrule
        MOTIP DETR & ResNet50\cite{he2016deep} & COCO & 32 & 50 & Step (at 20 and 35) & $2 \times 10^{-4} \;\big/\; 2 \times 10^{-6}$ & \revchanged{0.1} & 5.3\\
        MOTIP full & ResNet50 & PigDetect+PigTrack & 8 & 18 & Step (at 10 and 14) & $1 \times 10^{-4} \;\big/\; 1 \times 10^{-6}$ & \revchanged{0.1}  & 17.0\\
        MOTRv2 & ResNet50 & DanceTrack & 8 & 20 & Step (at 10) & $2 \times 10^{-4} \;\big/\; 2 \times 10^{-5}$ & \revchanged{0.1}  & 11.5\\
        \bottomrule
    \end{tabular}
\end{table*}

For MOTIP, we first pre-trained the DETR-based detection module on the PigDetect training and validation set and the PigTrack training set. The model achieving the best detection accuracy (see \secref{sec:track_eval_metrics}) on the PigTrack validation set was used as initialization for training the full model. MOTRv2 was initialized with a model pre-trained on DanceTrack \cite{sun2022dancetrack}. Bounding box priors for MOTRv2 were obtained from Co-DINO trained on the PigDetect training set, as described in \secref{sec:bench_ob_imp}. 

We evaluated three training configurations for both MOTIP and MOTRv2. In the default setup (1), the models were trained and validated according to the PigTrack dataset splits described in \secref{sec:pigtrack_splits}, and the best-performing model in terms of HOTA (see \secref{sec:track_eval_metrics}) on the validation set was used to obtain test performance. In a second setup (2), the validation set was added to the training data, with only two videos held out for validation. In a third setup (3), all test videos except one were also included in the training set, again retaining the same two videos for validation. By repeating this process for each of the 25 test videos, test performance can be evaluated in a leave-one-out fashion similar to cross-validation. In all three settings, we also included the PigDetect dataset during training by generating pseudo tracks from the images, following previous works \cite{zhou2020tracking, meinhardt2022trackformer, zhang_motrv2_2023}

To reduce the effect of convergence issues that are known to exist for both MOTRv2 and MOTIP, all training runs were conducted twice, and the model checkpoint with the best validation performance was selected to evaluate test performance. We used the official source code of MOTIP and MOTRv2 for training. The basic training configurations for both models are summarized in \tab{tab:training_spec_tr}.

For the SORT-based models (see \secref{sec:sort}), we also used Co-DINO trained on the PigDetect training set (see \secref{sec:bench_ob_imp}) as an object detector, \ch{without additional fine-tuning on the PigTrack dataset}. The hyperparameters of all SORT-based methods were tuned via Bayesian optimization \cite{snoek2012practical} using Weights and Biases \cite{wandb}, with the objective of maximizing HOTA on the PigTrack validation set. The best hyperparameter configurations identified through this optimization procedure were used to obtain test results. All SORT-based models were run with code from the BoxMOT repository (version 11.0.6 \cite{Brostrom_BoxMOT_pluggable_SOTA}).

\subsection{Evaluation metrics}\label{sec:track_eval_metrics}

For the evaluation of tracking, two aspects need to be considered: (1) The accuracy of object detections on individual video frames and (2) the accuracy of their associations across time. To assess these components, we employ the metrics introduced by \citet{luiten2021hota}. As for object detection evaluation (see \secref{sec:det_eval_metrics}), the first step to construct these metrics is to establish a matching between predicted and ground-truth bounding boxes and to enforce a minimum IoU of \(\alpha\) for all pairs. For a given \(\alpha\), the set of all matched pairs is referred to as true positives (\(\TP\)), the set of all unmatched predictions as false positives (\(\FP\)), and the set of all unmatched ground truths as false negatives (\(\FN\)). The detection accuracy is then simply defined as the proportion of \(\TP\):
\[
\deta = \frac{|\TP|}{|\TP| + |\FP| + |\FN|}
\]

\(\deta\) does not take into account the predicted object ID. That is, a \(\deta\) of 100\% is possible even if bounding boxes are incorrectly associated over time. To get an evaluation metric for association that is structurally equivalent to \(\deta\), the concept of true positive associations (\(\TPA\)), false positive associations (\(\FPA\)), and false negative associations (\(\FNA\)) has been proposed. Unlike \(\TP\), \(\FP\) and \(\FN\), which are calculated once for all predictions and ground truths in the dataset that is evaluated, \(\TPA\), \(\FPA\) and \(\FNA\) are calculated separately for each \(c\in \TP\). \(\TPAC\) comprises all \(\TP\) that have the same ground-truth object ID (gtID) and predicted object ID (prID) as \(c\):

\[
\TPAC = \{k\in\TP \mid \text{prID}(k) = \text{prID}(c) \land \text{gtID}(k) = \text{gtID}(c)\}
\]

\(\FPAC\) comprises (1) all \(\TP\) that have the same prID as \(c\) but a different gtID, and (2) all \(\FP\) that share the same prID as \(c\):

\begin{align*}
\FPAC &= \{k \in \TP \mid \text{prID}(k) = \text{prID}(c) \land \text{gtID}(k) \neq \text{gtID}(c)\} \\
      &\quad \cup \{k \in \FP \mid \text{prID}(k) = \text{prID}(c)\}.
\end{align*}

Finally, \(\FNAC\) comprises (1) all \(\TP\) that have the same gtID as \(c\) but a different predID, and (2) all \(\FN\) with the same gtID as \(c\):

\begin{align*}
\FNAC &= \{k \in \TP \mid \text{prID}(k) \neq \text{prID}(c) \land \text{gtID}(k) = \text{gtID}(c)\} \\
    &\quad \cup \{k \in \FN \mid \text{gtID}(k) = \text{gtID}(c)\}
\end{align*}

The proportion of \(\TPA\) is then calculated and averaged over all \(c\in \TP\) to obtain a measure of association accuracy:

\[
A(c) = \frac{|\TPAC|}{|\TPAC| + |\text{FNA}(c)| + |\text{FPA}(c)|}
\]
\[
\text{AssA}_\alpha = \frac{1}{|\TP|} \sum_{c \in \{\TP\}} A(c)
\]

The geometric average of \(\deta\) and \(\assa\) is called higher order tracking accuracy (\(\hota\)) and is an overall measure for tracking performance that takes into account both detection and association:

\[
\text{HOTA}_\alpha = \sqrt{\text{DetA}_\alpha \cdot \text{AssA}_\alpha}
\]

By averaging over different IoU detection thresholds \(\alpha\), we obtain the final metrics that are used for evaluating tracking methods in this paper:

\[
\text{HOTA}=\frac{1}{|A|} \sum_{\alpha \in A} \text{HOTA}_{\alpha}
\]

\[
\text{DetA} = \frac{1}{|A|} \sum_{\alpha \in A} \text{DetA}_{\alpha}
\]

\[
\text{AssA} = \frac{1}{|A|} \sum_{\alpha \in A} \text{AssA}_{\alpha}
\]
where \( A = \{0.05, 0.1, 0.15, \dots, 0.95\} \) is the set of IoU thresholds considered and \(|A|\) denotes the number of elements in the set.

For the sake of completeness, we also report MOTA \cite{bernardin2008evaluating}, which is also a measure for both detection and tracking performance, but has been criticized for exaggerating the importance of detection \cite{luiten2021hota}. Furthermore, we report the IDF1 score \cite{ristani2016performance}, which is an alternative measure for association performance. \revchanged{We refer to the original publications for detailed definitions of MOTA and IDF1, as a full explanation of all metrics is beyond the scope of this work.} Finally, we also report the number of ID switches (IDSW), i.e. the number of times where an object is assigned a different ID than in the previous frame. Note that IDSW should not be seen as a measure for association performance, as it ignores the duration of switches and penalizes cases where the ID switches back to the correct one.

\section{Results and Discussion}

\subsection{Object detection results on PigDetect}

\begin{figure*}[t]
    \centering
    \includegraphics[width=1\textwidth]{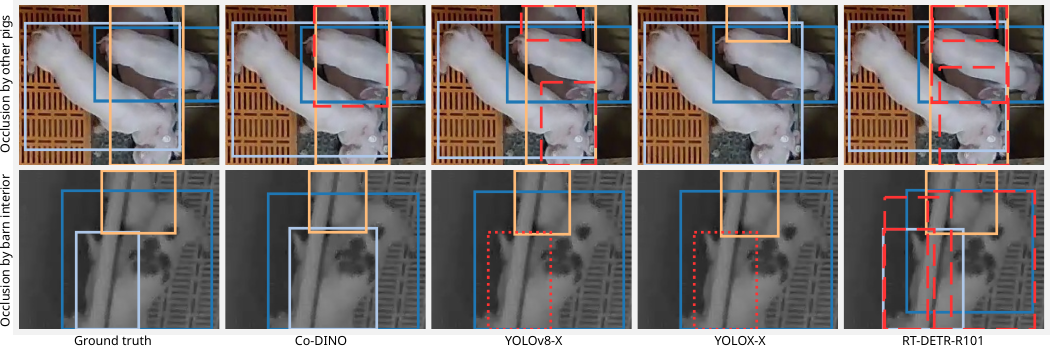}
    \caption{Visualization of challenging detection scenarios with model predictions. Shown are zoomed-in views of particularly difficult regions in two PigDetect test images. False positives are indicated by red dashed boxes, false negatives by red dotted boxes. All predicted bounding boxes with a score $\geq0.2$ are visualized, \revchanged{since low-confidence detections are frequently utilized in downstream tasks, for example in multi-object tracking \cite{zhang2022bytetrack, aharon2022bot}.}}
    \label{fig:qual_det}
\end{figure*}

Object detection performance on the PigDetect test set was benchmarked using YOLOX, YOLOv8 \revchanged{and RT-DETR, three} recent real-time models, and Co-DINO, a state-of-the-art model optimized for detection quality rather than inference speed (see \secref{sec:eval_od_methods}). Results for YOLOX, YOLOv8 \revchanged{and RT-DETR} were obtained for various model sizes. As expected, YOLOv8 outperforms YOLOX, reflecting the performance improvements associated with more recent real-time object detection models (\tab{tab:method_comparison}). \revchanged{In contrast to the results reported by \citet{zhao2024detrs} on the COCO dataset, RT-DETR performs worse than YOLOv8 on PigDetect and achieves results comparable to the older YOLOX model. Nonetheless, its speed-accuracy trade-off is more favorable. For example, RT-DETR-R50 matches YOLOX-L in accuracy while running twice as fast.} The results further suggest that additional gains can be achieved by using state-of-the-art models instead of real-time detectors. For example, an increase of roughly 2~\% AP can be observed when using Co-DINO, which is currently state of the art on the COCO object detection dataset, compared to YOLOv8-X. However, the inference speed of Co-DINO is considerably slower, which makes its usage in real-time applications more difficult. In the end, practitioners must decide which trade-off in terms of performance and inference speed is desirable for the application at hand. \revchanged{It should be noted that RT-DETR and Co-DINO might perform increasingly well with larger training corpora, as transformer-based architectures typically reach their full potential at scale.}

The consistent improvement in \(\text{AP}_{0.5}\) and AP when using increasingly powerful models suggests that PigDetect contains a sufficiently high number of challenging scenarios to differentiate between methods. This is an essential characteristic of a benchmark dataset. However, it should be noted that the error in terms of \(\text{AP}_{0.5}\) is relatively small for all benchmarked models, which indicates that most pigs in the test set can be roughly detected. This is facilitated by the top-view cameras used to capture the images. Furthermore, complex detection scenarios typically involve only a few pigs per image, sometimes even just a single one, while the majority of pigs is easily detectable. \fig{fig:qual_det} illustrates examples of challenging scenarios in the PigDetect dataset including heavy occlusions and poor visibility. In these cases, all models exhibit suboptimal performance, as indicated by false positive and false negative predictions as well as inaccurately predicted object boundaries. 

\begin{table}[t]
    \centering
    \caption{Performance measures for different detection methods on the PigDetect test set. Speed indicates inference time per image. Best results are in bold, second-best underlined.}
    \label{tab:method_comparison}
    \small
    \setlength{\tabcolsep}{3.5pt}
    \begin{tabularx}{\linewidth}{l c c >{\centering\arraybackslash}X c}
        \toprule
        Model & AP ↑ & \(\text{AP}_{0.5}\) ↑ & Speed (ms) ↓ & Parameters (M) \\
        \midrule
        YOLOX-S & 78.9 & 97.4 & 32& 8.9 \\
        YOLOX-M & 80.9 & 97.5 & 37& 25.3 \\
        YOLOX-L & 82.5 & 98.5 & 44& 54.1 \\
        YOLOX-X & 82.7 & 98.5 & 50& 99.0 \\
        \midrule
        YOLOv8-S & 80.3 & 97.8 & 30& 11.1 \\
        YOLOv8-M & 82.5 & 98.4 & 32& 25.9 \\
        YOLOv8-L & 83.2 & 98.6 & 40& 43.6 \\
        YOLOv8-X & \underline{83.6} & \underline{98.6} & 40 & 68.2 \\
        \midrule
        \revchanged{RT-DETR-R18} & \revchanged{80.8} & \revchanged{98.1} & \revchanged{\textbf{16}} & \revchanged{20.1} \\
        \revchanged{RT-DETR-R34} & \revchanged{80.6} & \revchanged{98.1} & \revchanged{\underline{17}} & \revchanged{31.4} \\
        \revchanged{RT-DETR-R50} & \revchanged{82.6} & \revchanged{98.3} & \revchanged{22} & \revchanged{42.9} \\
        \revchanged{RT-DETR-R101} & \revchanged{82.5} & \revchanged{98.4} & \revchanged{29} & \revchanged{76.6} \\
        \midrule
        Co-DINO & \textbf{85.5} & \textbf{98.7} & 111.2 &  234.7 \\
        \bottomrule
    \end{tabularx}
\end{table}

\subsection{\revchanged{Object detection results on third-party dataset}}
\revchanged{\citet{melfsen2023describing} introduced a pig object detection dataset comprising images from a single pen environment captured from above at an angle of roughly $45^{\circ}$. The test partition of this dataset comprises 120 images, on which we evaluated the performance of the models trained on PigDetect (\tab{tab:method_comparison_melfsen}). All models successfully detect the majority of pigs (\(\text{AP}_{0.5}\) ranging from 98.8 to 98.9). When taking into account localization accuracy, larger models generally achieve better performance, consistent with our observations on PigDetect. The best-performing models are RT-DETR-R50 and Co-DINO (AP~=~78.6). Overall, performance differences between methods are less pronounced than on PigDetect, suggesting that the dataset contains less challenging examples to clearly differentiate between methods.}

\revchanged{Interestingly, \citet{melfsen2023describing} reported a considerably higher performance using YOLOv7 (AP~=~86.4). Upon inspection of their training set, we found that this discrepancy is most likely due to near-identical training and test images (see \fig{fig:app_melfsen}), which make the results incomparable to ours. It appears that, to a large extent, the training and test sets were sampled from the same sequences without ensuring temporal disjointness.}

\revchanged{Beyond this evaluation, we also applied Co-DINO to an unseen video dataset to obtain detections for tracking. This resulted in substantially improved tracking performance compared to the method proposed by the authors of the video dataset (see \secref{sec:fto}). Examples of predicted bounding boxes on this video dataset, the PigDetect test set, and the test set of \citet{melfsen2023describing} are shown in \fig{fig:app_det_global}. These qualitative results further corroborate the fact that models trained on PigDetect can provide high-quality detections in previously unseen environments.}

\begin{table}[t]
    \centering
    \caption{\revchanged{Performance measures for different detection methods on the test set published by \citet{melfsen2023describing}. Best results are in bold, second-best underlined.}}
    \label{tab:method_comparison_melfsen}
    \small
    \setlength{\tabcolsep}{3.5pt}
    \begin{tabular}{l c c}
        \toprule
        Model & AP ↑ & \(\text{AP}_{0.5}\) \\
        \midrule
        \multicolumn{3}{c}{\textit{Trained on data from \cite{melfsen2023describing}}} \\
        YOLOv7-default & \textbf{86.4} & - \\
        \midrule
        \multicolumn{3}{c}{\textit{Trained on PigDetect}} \\
        YOLOX-S & 76.2 & 98.9 \\
        YOLOX-M & 77.2 & 98.9 \\
        YOLOX-L & 77.7 & 98.9 \\
        YOLOX-X & 77.7 & 98.9 \\
        \midrule
        YOLOv8-S & 76.4 & 98.8 \\
        YOLOv8-M & 77.6 & 98.9 \\
        YOLOv8-L & 77.7 & 98.9 \\
        YOLOv8-X & 77.6 & 98.9 \\
        \midrule
        RT-DETR-R18 & 76.6 & 98.9 \\
        RT-DETR-R34 & 77.5 & 98.9 \\
        RT-DETR-R50 & \underline{78.6} & 98.9 \\
        RT-DETR-R101 & 77.5 & 98.8 \\
        \midrule
        Co-DINO & \underline{78.6} & 98.9 \\
        \bottomrule
    \end{tabular}
\end{table}

\subsection{Challenging vs. random images: effects on detection}\label{res:chall}
\begin{figure}[t]
    \centering
    \includegraphics[width=\columnwidth]{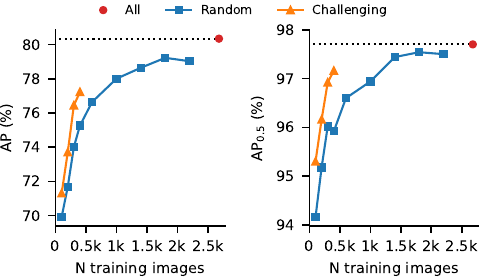}
    \caption{Experiment on the effect of including challenging images during training. For all experimental conditions, we report the average performance across five runs measured in AP (left) and \(\text{AP}_{0.5}\) (right). ``All'' indicates the performance when adding the challenging images on top of the random images for training. The black dotted line serves as a visual reference for this performance.}
    \label{fig:det_performance}
\end{figure}

To assess whether the targeted selection of challenging images has a positive effect on detection performance, we conducted an experiment where we systematically varied the two factors \enquote{number of training images} and \enquote{random vs.\ challenging} (see \secref{sec:met_cha}). The experiment yielded two main results: (1) Given the same number of training images, using challenging images leads to a better detection performance compared to using randomly selected images and (2) when only using randomly selected images, the detection performance tends to plateau, which can be counteracted by additionally including challenging images during training. These results are evident in \fig{fig:det_performance}, which depicts the detection performance for all experimental conditions.

With 400 images, the \(\text{AP}_{0.5}\) increased by more than 1~\% when using challenging (97.16~\%) compared to random images (95.92~\%). Even when using 1000 random images, the \(\text{AP}_{0.5}\) is still lower with 96.94~\%. When considering the AP, a similar pattern can be observed. This indicates that the information contained in the specifically selected challenging images is higher than in random images. Another observable trend is that the increase in performance through additional training images diminishes with larger sample sizes. For example, the performance gain in terms of \(\text{AP}_{0.5}\) is roughly 1~\% when using 200 vs.\ 100 random images. However, the difference between using 1400, 1800 or 2200 random images is negligible. This plateau effect indicates that randomly sampled images at some point contain little to no new information which could be learned by a detection model. This is in line with our observation that hard detection cases are rare and need to be specifically identified. In terms of AP, performance increases by more than 1~\% when using all images (80.34~\%) compared to using 1800 (79.22~\%) or 2200 random images (79.02~\%). Even for the \(\text{AP}_{0.5}\), a small increase can be observed, which indicates that some edge cases which have not been detected before can be detected through the inclusion of challenging training images.

\subsection{Multi-object tracking results on PigTrack}\label{sec:res_track}

\begin{table}[t]
    \centering
    \caption{Performance measures for different tracking methods on the PigTrack test set. Best results are in bold, second-best underlined. Results for MOTRv2 and MOTIP are shown for three training data setups (see \secref{sec:track_implementation}): (1) \textit{train} = train set only, (2) \textit{+val} = train + val set, and (3) \textit{LOO} = train + val + test set in leave-one-out fashion.}
    \label{tab:tracking_metrics}
    \footnotesize
    \setlength{\tabcolsep}{2.5pt}
    \begin{tabularx}{\linewidth}{l c c c c c c}
        \toprule
        Method & HOTA ↑ & DetA ↑ & AssA ↑ & MOTA ↑ & IDF1 ↑ & IDSW ↓\\
        \midrule
        ByteTrack    & 80.6 & 81.5 & 80.2 & 93.9 & 90.7 & 206\\
        StrongSORT   & 83.8 & 87.0 & 80.9 & 94.0 & 87.8 & 672 \\
        OC-SORT      & 86.8 & 88.4 & 85.4 & 95.3 & 91.4 & 277\\
        Deep OC-SORT & 86.9 & 88.1 & 85.9 & 94.7 & 91.6 & 286\\
        ImprAsso     & 87.3 & \textbf{88.7} & 86.1 & \underline{95.9} & 92.1 & 238\\
        BoT-SORT     & \underline{87.9} & \underline{88.5} & 87.4 & \textbf{96.1} & 93.5 & 165\\
        \textcolor{gray!60}{BoT-SORT (overfit)} & \textcolor{gray!60}{88.4} & \textcolor{gray!60}{88.5} & \textcolor{gray!60}{88.5} & \textcolor{gray!60}{96.2} & \textcolor{gray!60}{94.5}  & \textcolor{gray!60}{134}\\

        \midrule
        MOTRv2 (train)       & 87.6 & 87.4 & 88.0 & 94.3 & 93.3  & 204\\
        MOTRv2 (+val)      & \textbf{88.2} & 87.9 & \textbf{88.7} & 94.5 & \underline{93.9}  & 221\\
        MOTRv2 (LOO)       & 87.4 & 87.4 & 87.5 & 93.8  & 92.2  & 414\\
        \midrule
        MOTIP (train)        & 83.5 & 84.6 &  82.6 & 91.2 & 89.7  & 3538\\
        MOTIP (+val)        & 86.5 & 85.4 & \underline{87.9} & 93.2  & \textbf{94.2}  & 1086\\
        MOTIP (LOO)        & 86.7 & 85.8 & 87.8 & 93.3 & 93.7  & 584\\
        \bottomrule
    \end{tabularx}
\end{table}

Multi-object tracking performance on the PigTrack test set was benchmarked with several recent SORT-based models and the two end-to-end trainable trackers MOTRv2 and MOTIP (see \secref{sec:eval_mot}). When comparing the two model classes, the main finding is that recent SORT-based models achieve superior detection performance, while end-to-end models demonstrate better association performance (\tab{tab:tracking_metrics}). 

For example, both ImprAssoc (DetA=88.7, MOTA=95.9) and BoT-SORT (DetA=88.5, MOTA=96.1) outperform the best training runs of MOTRv2 (DetA=87.9, MOTA=94.5) and MOTIP (DetA=85.8, MOTA=93.3) on metrics that are exclusively (DetA) or primarily (MOTA) driven by detection performance. This difference is due to the fact that end-to-end models rely on joint architectures for detection and tracking that are not optimized for detection performance. In contrast, SORT-based methods are able to leverage carefully engineered external object detectors such as Co-DINO, which we employ in our work (see \secref{sec:codino}). The large DetA difference between the two end-to-end methods can be attributed to MOTRv2 also leveraging Co-DINO bounding boxes as detection priors (see \secref{sec:track_implementation}), whereas MOTIP relies solely on its integrated deformable DETR detector.

Regarding association performance, the best training runs of MOTRv2 (AssA=88.7, IDF1=93.9) and MOTIP (AssA=87.9, IDF1=94.2) achieve a superior performance compared to the best SORT-based methods, ImprAsso (AssA=86.1, IDF1=92.1) and BoT-SORT (AssA=87.4, IDF1=93.5). This finding is in line with previous results on the DanceTrack dataset \cite{sun2022dancetrack}, which shares key characteristics with PigTrack, such as visually similar individuals, non-linear motion, and frequent occlusions. Under such conditions, learned association strategies appear to outperform the association heuristics commonly employed by SORT-based methods. Even when overfitting the hyperparameters of BoT-SORT to the test set (AssA=88.5, IDF1=94.5), it fails to consistently surpass the association performance of learned trackers, indicating inherent limitations of heuristic-based association strategies. The performance of the best overfitted model was included, as models in practice are often intended to operate in specific environments where overfitting is desirable.

When taking into account both detection and association, there are no notable differences between the best-performing SORT-based and end-to-end methods. BoT-SORT achieves a HOTA of 87.2, while MOTRv2 achieves a HOTA of 88.2. 

We also investigated whether the performance of the end-to-end trackers can be improved by adding more in-domain training data. Surprisingly, our results indicate that this cannot be consistently achieved across the range of data available to us. While the HOTA of MOTRv2 and MOTIP increases by 0.6 and 3.0, respectively, when the validation set is added to the training data, no notable improvement is observed when additionally incorporating the test set in a leave-one-out fashion. In fact, the HOTA of MOTRv2 even decreases slightly by 0.2 compared to the baseline trained on the training set only. This finding is surprising, as the test set was manually selected to include particularly challenging tracking scenarios that would be expected to provide valuable information during training. Further research is required to identify the exact reasons for this phenomenon. \revchanged{Determining the optimal amount and composition of training data for end-to-end trackers appears to be a non-trivial task.}

\subsection{Multi-object tracking results \revchanged{on third-party dataset}}\label{sec:fto}
\begin{table}[t]
    \centering
    \caption{Performance measures for different tracking methods on the dataset introduced by \citet{yu2025fto}. Best results are in bold, second-best underlined. The first block contains results taken from \citet{yu2025fto}, while all other blocks contain results of the models from this work.}
    \label{tab:tracking_metrics_fto}
    \footnotesize
    \setlength{\tabcolsep}{2.5pt}
    \begin{tabularx}{\linewidth}{l c c c c c c}
        \toprule
        Method & HOTA ↑ & DetA ↑ & AssA ↑ & MOTA ↑ & IDF1 ↑ & IDSW ↓\\
        \midrule
        \multicolumn{7}{c}{\textit{Using YOLOv8 trained on data from \cite{riekert_automatically_2020}}} \\
        BoT-SORT & 66.3 & - & - & 78.5 & 75.1 & - \\
        FTO-SORT & 75.6 & - & - & 83.1 & 90.1 & - \\
        \midrule
        \multicolumn{7}{c}{\textit{Using YOLOv8 trained on PigDetect}} \\
        ByteTrack       & 77.0 & 76.7 & 77.7 & 90.4 & 88.5 & 60 \\
        StrongSORT      & 75.1 & 77.6 & 72.9 & 89.0 & 83.4 & 94 \\
        OC-SORT         & 76.1 & 79.6 & 73.0 & 91.1 & 83.9 & 80 \\
        Deep OC-SORT    & 76.3 & 77.6 & 75.2 & 88.8 & 85.1 & 78 \\
        ImprAsso        & 77.4 & 78.0 & 77.0 & 90.2 & 87.2 & 52 \\
        BoT-SORT        & 81.2 & 79.3 & 83.3 & 91.1 & 93.5 & 9 \\
        \midrule
        \multicolumn{7}{c}{\textit{Using Co-DINO trained on PigDetect}} \\
        ByteTrack       & 83.2 & 81.3 & \underline{85.4} & \underline{95.8} & \underline{97.4} & 8 \\
        StrongSORT      & 79.9 & \underline{82.2} & 77.9 & 94.9 & 88.5 & 88 \\
        OC-SORT         & 82.6 & 81.5 & 83.9 & \underline{95.8} & 95.3 & 15 \\
        Deep OC-SORT    & 81.9 & 80.8 & 83.3 & 94.8 & 93.9 & 17 \\
        ImprAsso        & \underline{83.6} & 82.1 & \underline{85.4} & 95.4 & 96.8 & \textbf{3} \\
        BoT-SORT        & \textbf{84.9} & \textbf{83.2} & \textbf{86.8} & \textbf{96.4} & \textbf{97.9} & \underline{4} \\
        \midrule
        \multicolumn{7}{c}{\textit{Trained on PigDetect and PigTrack}} \\
        MOTRv2 (+val)   & 81.1 & 80.2 & 82.3 & 92.0 & 92.7 & 18 \\
        MOTIP (+val)    & 79.0 & 77.4 & 80.9 & 90.7 & 92.4 & 46 \\
        \bottomrule
    \end{tabularx}
\end{table}

Recently, \citet{yu2025fto} presented FTO-SORT, a tracking system based on YOLOv8 \cite{Jocher_Ultralytics_YOLO_2023} and BoT-SORT \citep{aharon2022bot}, designed to enhance generalization performance through several modifications. During object detection training, they employ custom data augmentations and a loss function aimed at improving the detection of challenging pig instances. For multi-object tracking, they incorporate a specialized Re-ID module, Farm Track-id Optimizer (FTO), into BoT-SORT. They trained their detector on data published by \citet{riekert_automatically_2020} and evaluated tracking performance on a 2.5~min test video from a completely different pen setting (details on datasets in \tab{tab:pigtrack}). The results indicate that the baseline (default YOLOv8 with BoT-SORT) can be vastly improved by adding the modifications of FTO-SORT. For example, the best-performing FTO-SORT model achieves a HOTA of 75.6, while the baseline only achieves a HOTA of 66.3 (\tab{tab:tracking_metrics_fto}, first block).

To evaluate the out-of-the-box performance of the tracking models from this work in an entirely new pen setting, we also applied them on the dataset introduced by \citet{yu2025fto} without any additional fine-tuning or hyperparameters adjustments. For a fair comparison to FTO-SORT, we evaluated the tracking models not only with the more powerful Co-DINO detector, but also with YOLOv8 trained on PigDetect, as described in \secref{sec:bench_ob_imp}. Looking at the performance of our YOLOv8-based trackers (\tab{tab:tracking_metrics_fto}, second block), we find that all except StrongSORT outperform FTO-SORT. For example, BoT-SORT improves upon FTO-SORT by 4.6 HOTA points. When using the same models with the more powerful Co-DINO detector (\tab{tab:tracking_metrics_fto}, third block), the difference becomes even more pronounced. For example, BoT-SORT surpasses FTO-SORT by 9.3 HOTA points. End-to-end models (\tab{tab:tracking_metrics_fto}, fourth block) perform worse than SORT-based methods on this new dataset. This contrasts with the results reported in \secref{sec:res_track} and suggests poorer generalization to unseen pen settings. Only when using the SORT-based models with a weaker detector (YOLOv8), the end-to-end methods can compete with them. Nonetheless, both end-to-end models still outperform FTO-SORT.

The results obtained on the dataset of \citet{yu2025fto} convey an important insight: To obtain powerful and generalizable models in practice, a lot can already be achieved by training and hyperparameter-tuning general-purpose models on carefully curated, diverse datasets. While further improvements via incremental architecture changes or other modifications are certainly possible, we argue that establishing strong baselines in a data-centric manner is a crucial first step for all further analyses and comparisons.

\subsection{Multi-object tracking oracle analysis on PigTrack}
\begin{table}[t]
    \centering
    \caption{Oracle performance of SORT-based methods on the PigTrack test set.}
    \label{tab:oracle_tracking_metrics}
    \footnotesize
    \setlength{\tabcolsep}{3.3pt}
    \begin{tabularx}{\linewidth}{l c c c c c c}
        \toprule
        Method & HOTA ↑ & DetA ↑ & AssA ↑ & MOTA ↑ & IDF1 ↑ & IDSW ↓\\
        \midrule
        ByteTrack    & 89.4      &  88.7     & 90.2      & 98.4      &  97.8    & 45\\
        StrongSORT &  98.7 &  99.4 &  98.0 &  99.4 & 98.9  & 32\\
        OC-SORT      & 98.7      & 100.0      & 97.5      & 99.9      & 98.4      &77\\
        Deep OC-SORT &98.6  & 99.9 & 97.4 &  99.9 & 98.1     &  66\\
        ImprAsso   & 98.9 &  99.9 & 98.0 & 100.0 & 98.9     &  31 \\
        BoT-SORT  &  97.9 &  98.4 & 97.4 & 100.0 & 99.1    &   39\\
        \bottomrule
    \end{tabularx}
\end{table}

To assess the association performance of SORT-based trackers in the ideal scenario that pigs are detected perfectly, we performed an oracle analysis in which all trackers were provided with ground-truth bounding boxes as input. In this setup, only the association problem remains to be solved. We evaluated the same SORT-based methods and hyperparameter configurations that were used to obtain the results in \secref{sec:res_track}.

For all methods, association performance on the PigTrack test set improves substantially when oracle detections are provided (\tab{tab:oracle_tracking_metrics}). For instance, both StrongSORT and BoT-SORT achieve an AssA of 98.0 and an IDF1 of 98.9. Similarly strong association performance has been reported for oracle analyses on the MOT17 pedestrian dataset \cite{milan2016mot16}, as shown by \citet{sun2022dancetrack}, suggesting that detection performance is often a bottleneck for SORT-based methods. These analyses suggest that SORT-based methods still have strong potential in case that object detection can be substantially improved. Whether this potential can be fully realized in practice using image-based detectors remains questionable, especially under challenging conditions such as extreme occlusions, poor visibility or motion blur, which are prevalent in PigTrack. The results for MOTRv2 and MOTIP (see \secref{sec:res_track}) further suggest that superior association performance can be achieved despite inferior detections. This suggests that good detections are not necessarily a prerequisite for good association performance.

\subsection{Error analysis of end-to-end trainable tracking models}
\begin{figure}[t]
    \centering
    \includegraphics[width=\columnwidth]{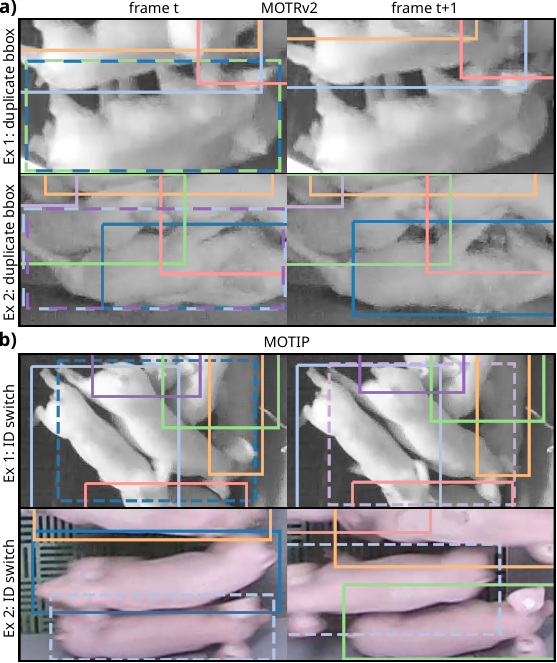}
    \caption{Characteristic failure cases of end-to-end trainable trackers. In both subfigures, the predicted IDs in frame t and frame t+1 are indicated by the color of the bounding boxes. If the predicted ID is the same in both frames, the bounding box has the same color. (a) shows two examples for MOTRv2. The two-colored dotted bounding boxes in frame t represent overlapping duplicate predictions made by the model in case of occlusions. \revchanged{In other words, in these cases the model produces two distinct bounding boxes with different predicted IDs at the same location.} In frame t+1, the corresponding two IDs are not predicted at all. (b) shows two examples for MOTIP. The dotted bounding boxes indicate where errors happen. For instance, in the top row, the central pig is incorrectly assigned a different ID in frame t+1 \revchanged{even though it barely moved.}}
    \label{fig:track_qual}
\end{figure}

\begin{figure}[t]
    \centering
    \includegraphics[width=\columnwidth]{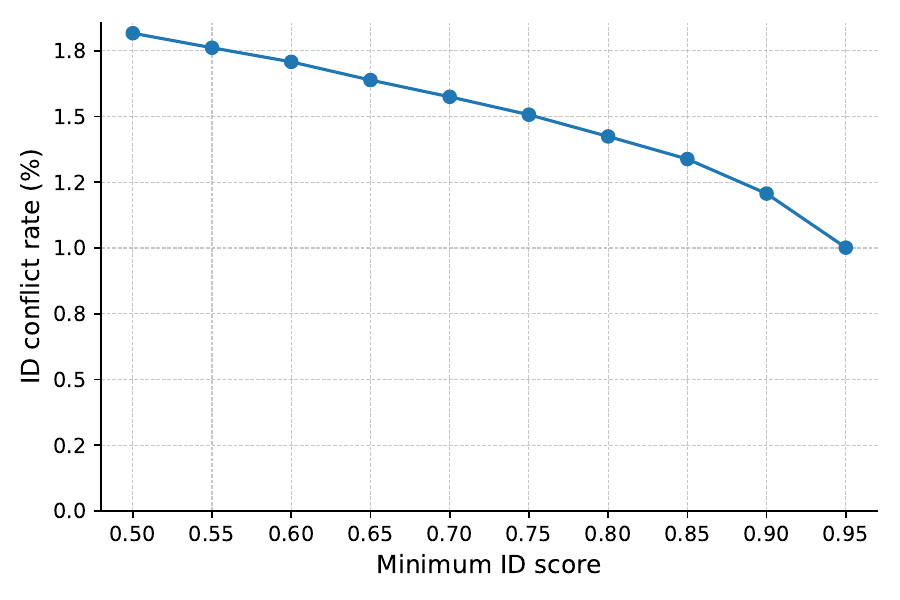}
    \caption{ID conflict rate in MOTIP for different minimum required ID scores.}
    \label{fig:latch_failure}
\end{figure}

The characteristics of SORT-based models have been studies for years, providing valuable insights for potential method improvements. In contrast, comparably little empirical evidence exists for the specific modes of failure of end-to-end trainable models. To reduce this gap, we report characteristic failure cases of MOTRv2 and MOTIP on the PigTrack test set. Interestingly, both methods tend to commit errors in scenarios where classical SORT-based methods are not expected to fail in a similar way, often leading to odd-looking tracking results.

MOTRv2 uses track queries to predict bounding boxes for each frame of the input video, where distinct objects are represented by distinct track queries (see \secref{sec:learning_based}). A common failure case emerges when one tracked objects severely occludes another one. In this case, the track query of the occluded object often latches onto the foreground object instead of not making any prediction. As a result, both track queries often produce near-identical bounding boxes (\fig{fig:track_qual}a, frame t). Following the duplicate predictions, we often observe on the subsequent frames that neither of the two track queries latches onto an object anymore, leading to odd-looking cases where pigs are not detected at all although they are clearly visible (\fig{fig:track_qual}a, frame t+1). We conjecture that this is due to the fact that duplicate predictions of the same object are penalized during training. If duplicate predictions still happen at inference time, this information can be retrieved via self-attention between track queries, and then be used to reduce the score of the affected bounding boxes. Similar errors have also been reported on the MOT17 pedestrian dataset \cite{milan2016mot16} by \citet{zhang_motrv2_2023}. Interestingly, they found that this issue does not occur on the DanceTrack dataset \cite{sun2022dancetrack} and attributed this to the benefits of a larger training set. MOT17 has \num[group-digits=integer]{5316} frames in the training set, while DanceTrack has \num[group-digits=integer]{41796}. In our study, increasing the training set size from \num[group-digits=integer]{9959} to roughly \num[group-digits=integer]{26000} frames (see \secref{sec:res_track}) did not consistently mitigate the issue.

MOTIP uses the ID predictions of bounding boxes on previous frames as information to predict IDs for all bounding boxes on the current frame (see \secref{sec:learning_based}). Surprisingly, incorrect ID predictions often emerge in simple scenarios, where classical association heuristics employed by SORT-based models would not produce any errors. For example, even when the bounding boxes on subsequent frames overlap almost perfectly (\fig{fig:track_qual}b top) or the pigs move in a simple linear fashion (\fig{fig:track_qual}b bottom), IDs might be predicted incorrectly. One potential reason for these kinds of failures are ID conflicts due to duplicate ID predictions. To count the number of ID conflicts, we identified for each ID the number of bounding boxes that predict this ID with a minimum score of \( s_{\min} \) even though another bounding box also predicts it with \( s_{\min} \). From this number, we then subtract one since an ID is supposed to be predicted one time. For example, if three bounding boxes predict the ID 15 with a score of at least \( s_{\min} \), this would be counted as two ID conflicts. \revchanged{The minimum ID score \( s_{\min} \) is a variable, which we varied from 0.5 to 0.95 in steps of 0.05. By running this analysis on the entire PigTrack dataset, the proportion of ID conflicts can be calculated for the different minimum ID scores \( s_{\min} \) (\fig{fig:latch_failure}).} In total, roughly 1.8\% of bounding boxes predict an ID that has already been assigned to another box with a score of at least 0.5. Even if a score of 0.95 is required for an ID conflict, roughly 1\% of all bounding boxes fulfil this criterion. Given that a pen in our dataset contains between 7 and 26 pigs, this means that such an error occurs on average every couple of frames. Since IDs can only be assigned uniquely, those duplicate ID predictions with the lowest confidence will in most cases get assigned a new ID, leading to the errors that are observed in \fig{fig:track_qual}b. Similarly, the large amount of ID switches observed for MOTIP (\tab{tab:tracking_metrics}) can be explained by this. Interestingly, these characteristic errors of MOTIP were not reported on the DanceTrack dataset by \citet{gao2025multiple}, and we were able to reproduce this finding. One possible explanation for this discrepancy is the abundance of in-domain data in the pretraining corpus: DanceTrack features videos of humans, and COCO includes a large number of images with humans. In contrast, COCO does not contain images with pigs. Additionally, DanceTrack is larger then PigTrack, potentially enabling more effective training, and has a higher frame rate, which may help reduce errors at inference time.

SORT-based models do not exhibit the characteristic errors of MOTRv2 and MOTIP since they simply use bounding boxes from a pre-trained tracker and associate them using motion and appearance heuristics. This makes it even more notable that MOTRv2 and MOTIP achieve a better association performance on PigTrack. Engineering efforts that tackle these issues might lead to improvements that further increase the gap in association performance between SORT-based and end-to-end trackers.

\section{Conclusion}
%%%%%%%%%%%%%%%%%%%%%%%%% BENCHMARK CONSTRUCTION + METHOD EVALUATION
In this work, we curated two benchmark datasets \ch{based on footage from conventional indoor barn environments: PigDetect for object detection and PigTrack for multi-object tracking. We then investigated the performance of several recent general-purpose methods on both datasets.} To our knowledge, this is the first work in the context of pig farming that focuses on the systematic acquisition of diverse and challenging benchmark datasets for detection and tracking. Such datasets are crucial prerequisites for method development and comparison in animal farming. Our benchmarking results indicate that the datasets are able to differentiate between the evaluated methods, often highlighting particular strengths and weaknesses, which is an essential characteristic of a benchmark dataset. 

Regarding object detection, substantial performance gains can be achieved by employing more optimized state-of-the-art models instead of real-time detectors, at the cost of slower inference. In the end, practitioners need to select the right model based on the desired performance-speed trade-off for the application at hand. Regarding multi-object tracking, we found that SORT-based models tend to achieve a better detection performance, while end-to-end trackers are better at association. Although better association performance is arguably more important than detection quality, end-to-end models come with additional drawbacks: Both annotation and training require large amounts of resources and the generalization to new pens is worse than for SORT-based models. For this reason, we do not currently consider them to be preferable to SORT-based methods. However, with the development of new end-to-end tracking methods in the future, the benefits might eventually outweigh their drawbacks. Given the relatively limited visual variability between different pigs and pen settings, the emergence of robust and broadly applicable end-to-end or SORT-based pig tracking models appears to be a realistic prospect. In fact, our trained detection models and the trackers based on it already achieved good performance in unseen pen settings. The benchmark datasets developed in this work can serve as a tool to constantly monitor new developments and identify models that improve upon the approaches that are currently dominating.

%%%%%%%%%%%%%%%%%%%%%%%%% CHARACTERISTICS OF BENCHMARK
We also showed experimentally that the inclusion of challenging images has a positive influence on detection performance. While we only investigated this effect for object detection, similar effects might also exist for other tasks such as multi-object tracking or action understanding. This highlights the importance of targeted selection strategies during annotation to obtain information-rich datasets. We conjecture that this is even more important in monotonous settings, of which pens in animal farming are a prime example. A careful selection of diverse and challenging training data might in many cases be the most effective way to reduce the error rate and obtain strong models for practical applications.

%%%%%%%%%%%%%%%%%%%%%%%%% LIMITATIONS
\revchanged{While the models trained on our datasets demonstrate promising results in unseen pen settings, several limitations must be acknowledged. (1) Although our datasets contain many challenging examples, this does not imply flawless performance in such cases. Scenarios like poor visibility or heavily overlapping animals might still lead to model errors, even in pen environments similar to those used during training. (2) The datasets that were used to evaluate test performance are limited in size and restricted to conditions that are relatively similar to the training data. Specifically, the majority of images and videos in our datasets originate from conventional indoor pens with common crossbreeds. Therefore, models might perform worse for outdoor pens, pens with unusual interiors or pig breeds, or substantially different camera perspectives. Future work should focus on combining and extending existing datasets to build a more comprehensive benchmark, and enable a more robust assessment of model performance across diverse settings. (3) Examples involving heavy piling and occlusions, where unambiguous annotation was not possible, were not included in our datasets. This illustrates the practical limitations of detection and tracking and highlights the need for further research on how to address such situations when using computer vision for individualized behavioral analysis. For example, it might be possible to leverage ear tag detection methods to recover lost tracks resulting from complex scenarios that current tracking models cannot handle. Re-identification via ear tags might also be useful in situations where a single camera cannot cover the entire pen.}

%%%%%%%%%%%%%%%%%%%%%%%%% OPEN DATA/ FUTURE WORK / DOWN STREAM APPLICATIONS
Going forward, the detection and tracking functionality introduced in this work can be used as a modular component in more sophisticated action understanding frameworks. For example, the capabilities of vision-language models \cite{you2023ferret, zhang2024ferret} that can be prompted with spatial localizations, such as bounding boxes, should be continuously monitored. These models might offer a convenient way to extract action-related information from images or videos in the future. Similarly, trainable models for spatio-temporal action understanding \cite{gritsenko_end--end_2023-1, ryali2023hiera, wu_memvit_2022} have not been thoroughly explored yet. For this purpose, our datasets might also be extended with spatio-temporal action labels. Since challenging tracking scenarios naturally include interesting actions such as mounting, playing or fighting, PigTrack is well suited for extending annotations to the action domain. All trained models, the source code for training and inference, as well as the PigDetect and PigTrack datasets are made publicly available. This way, researchers and practitioners can reproduce our results, use our models for downstream analyses, or add their own training data to obtain customized models for their setting. We aim to encourage open data and open source principles to enable a more holistic scientific process in the context of animal farming.

\section*{Code and data availability}
All research code including training, evaluation and inference functionality for the detection and tracking models from this paper are made publicly available at \url{https://github.com/jonaden94/PigBench}. The PigDetect dataset and trained pig detection models are available from \url{https://doi.org/10.25625/I6UYE9 }. The PigTrack dataset as well as trained pig tracking models can be found at \url{https://doi.org/10.25625/P7VQTP  }.

\section*{Acknowledgements}
This publication was funded with NextGenerationEU funds from the European Union by the Federal Ministry of Research, Technology and Space under the funding code 16DKWN038. The responsibility for the content of this publication lies with the authors.

The authors gratefully acknowledge the computing time granted by the Resource Allocation Board and provided on the supercomputer Emmy/Grete at NHR-Nord@Göttingen as part of the NHR infrastructure. The calculations for this research were conducted with computing resources under the project nib00034.

\section*{CRediT authorship contribution statement}
\textbf{J.~H.}: conceptualization, methodology, formal analysis, visualization, software, data curation, writing – original draft; 
\textbf{C.~P.}: conceptualization, methodology, data curation, writing – review and editing; 
\textbf{M.~Z.}: software; 
\textbf{P.~S.}: data curation; 
\textbf{E.~C.}: data curation; 
\textbf{A.~M.~Y.}: data curation;
\textbf{R.~Y.}: writing – review and editing; 
\textbf{T.~K.}: supervision, writing – review and editing; 
\textbf{I.~T.}: supervision, writing – review and editing.

\section*{Declaration of Competing Interest}

The authors declare that they have no known competing financial interests or personal relationships that could have appeared to influence the work reported in this paper.

\section*{Declaration of Generative AI and AI-assisted technologies in the writing process}

During the preparation of this work, the authors used ChatGPT \cite{OpenAI2023} in order to paraphrase text and generate basic code for data processing as well as figure and table generation. After using ChatGPT, the authors reviewed and edited the content as needed and take full responsibility for the content of the publication.

\cleardoublepage
\appendix
\section[\appendixname~\thesection]{Detection datasets} \label{app:sec_det}

\begin{table}[!h]
  \centering
  \small
  \caption{Dataset characteristics by image source for PigDetect.}
  \label{tab:source_stats}

  \begin{threeparttable}
    \setlength{\tabcolsep}{4pt}
    \begin{tabular}{@{}lrrrrrrr@{\hspace{.8em}}}
      \toprule
      Source & min\_res & max\_res & \#Images & \#Night &
              \#BBoxes & \#Challenging & \#Pens \\
      \midrule
      Alameer \cite{alameer_automated_2022}   &  630×353     &  811×509     &   308 & 153 &  3278 &  75 & 2\tnote{a} \\
      Bergamini \cite{bergamini_extracting_2021} & 1280×720   & 1280×720   &   186 &   0 &  1801 &  36 & 1 \\
      Psota \cite{psota_multi-pig_2019}     & 1920×1080 & 3840×2160 &  733 &  91 & \num[group-digits=integer]{10301} & 141 & 17 \\
      Henrich (our)    &  640×360     & 1920×1080 & 1704 & 443 & \num[group-digits=integer]{26797} & 316 & 11 \\
      \bottomrule
    \end{tabular}

    \begin{tablenotes}[para,flushleft]  % note stays inside table width
      \footnotesize
      \item[a] There are actually more than two pens; the original
               publication does not state an exact number, but only two
               visually distinct pen environments are present, which is why we put this number there. \\
    \end{tablenotes}
  \end{threeparttable}
\end{table}

\begin{figure*}[!b]        % ! = override constraints, h = right here
  \centering
  \includegraphics[width=\textwidth]{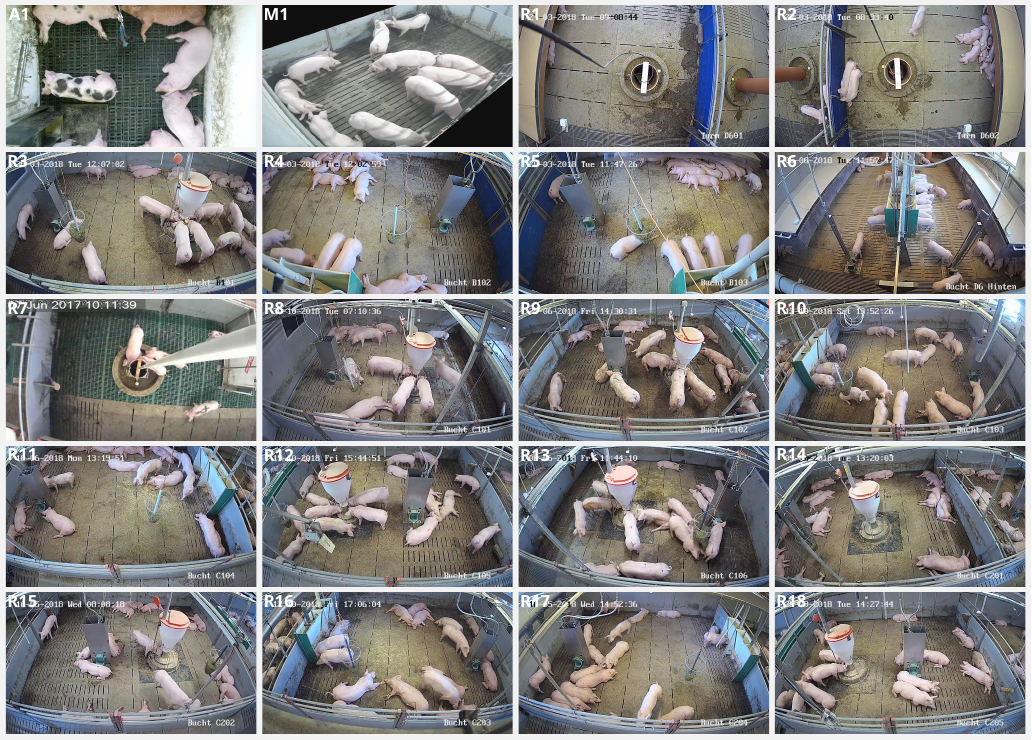}
  \caption{Example images from all unique pen environments contained in the publicly available annotated object detection datasets presented in \secref{sec:pig_cha}. The character in the top left corner of each image indicates the initial of the first author of the corresponding work (A\cite{alameer_automated_2020}, M\cite{melfsen2023describing}, R\cite{riekert_automatically_2020, riekert2021model}). The unique pen environments presented in each work are then numbered starting from one.}
  \label{fig:app_det_third}
\end{figure*}

\begin{figure*}[t]
    \centering
    \includegraphics[width=0.85\textwidth]{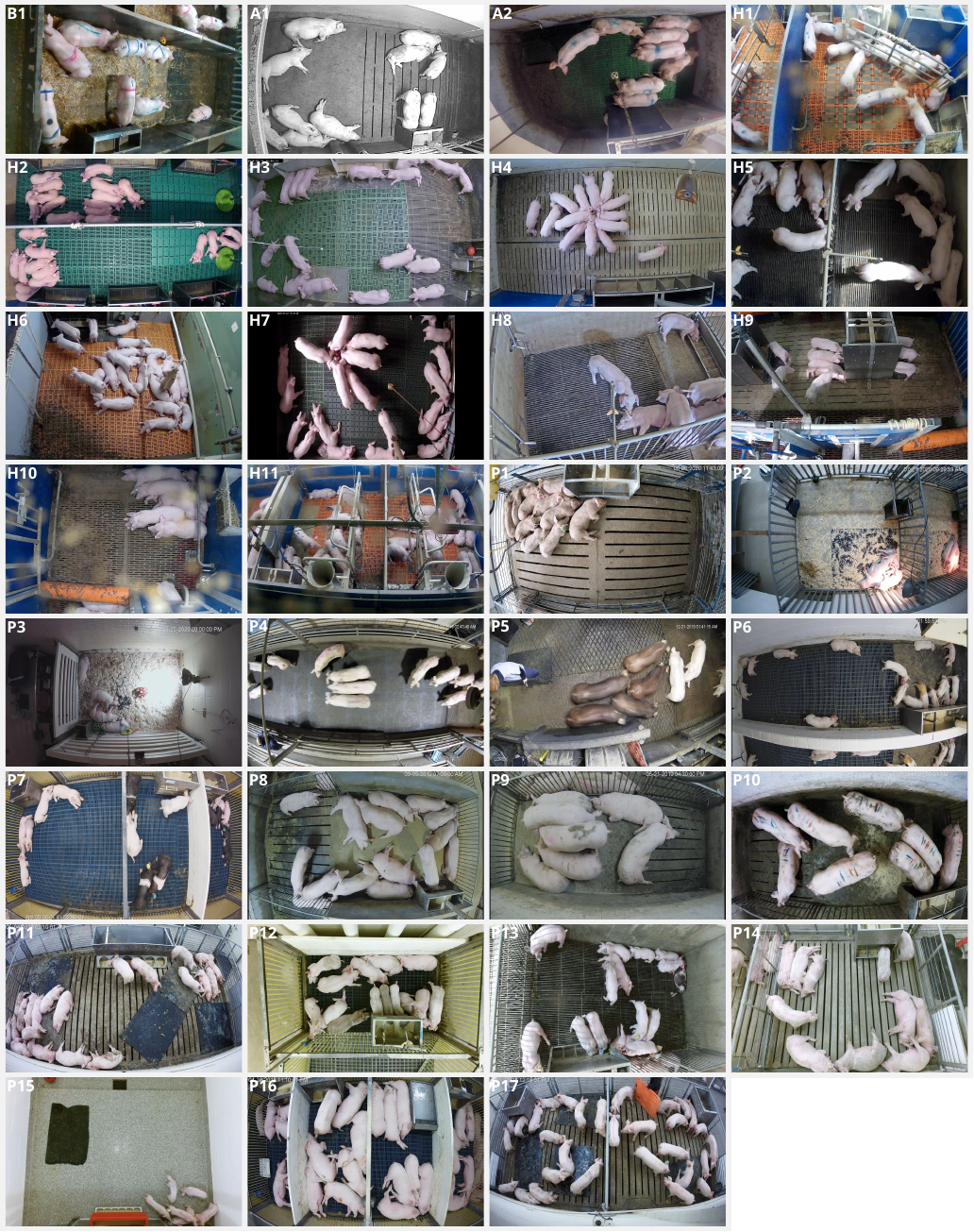}
    \caption{Example images from all unique pen environments contained in the PigDetect dataset. The character in the top left corner of each image indicates the initial of the first author of the work in which the original image data was released (B\cite{bergamini_extracting_2021}, A\cite{alameer_automated_2022}, P\cite{psota_multi-pig_2019}). H denotes Henrich, representing data introduced in this work. The unique pen environments from each source are then numbered starting from one. H6 is the pen environment of the PigDetect test set. Apart from different pen environments, PigDetect includes images from different camera angles. These are not depicted here.}
    \label{fig:app_det_ours}
\end{figure*}

\begin{figure*}[t]
    \centering
    \includegraphics[width=1\textwidth]{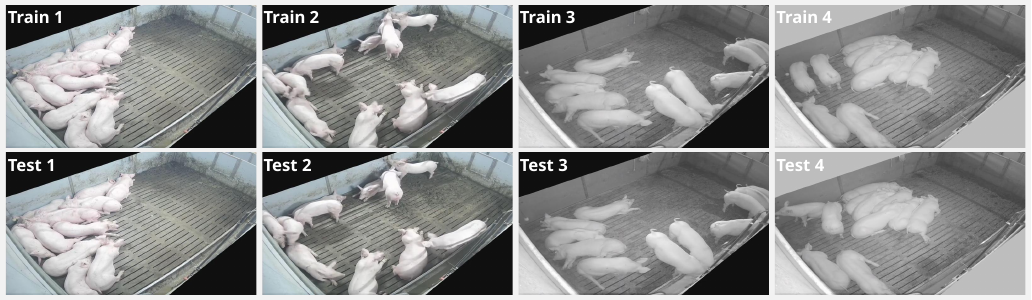}
    \caption{\revchanged{Examples for near-identical training and test images in the detection dataset published by \citet{melfsen2023describing}. For reference, we also provide the names of the respective images in the published dataset: Train 1 (954-0000-0001\_400.jpg), Train 2 (836-0125-0126\_750.jpg), Train 3 (931-0035-0037\_3150.jpg), Train 4 (ch01\_20211120121240.mp4\_10.jpg), Test 1 (954-0000-0001\_450.jpg), Test 2 (836-0125-0126\_800.jpg), Test 3 (931-0035-0037\_3100.jpg), Test 4 (ch01\_20211120121240.mp4\_9.jpg).}}
    \label{fig:app_melfsen}
\end{figure*}

\begin{figure*}[t]
    \centering
    \includegraphics[width=1\textwidth]{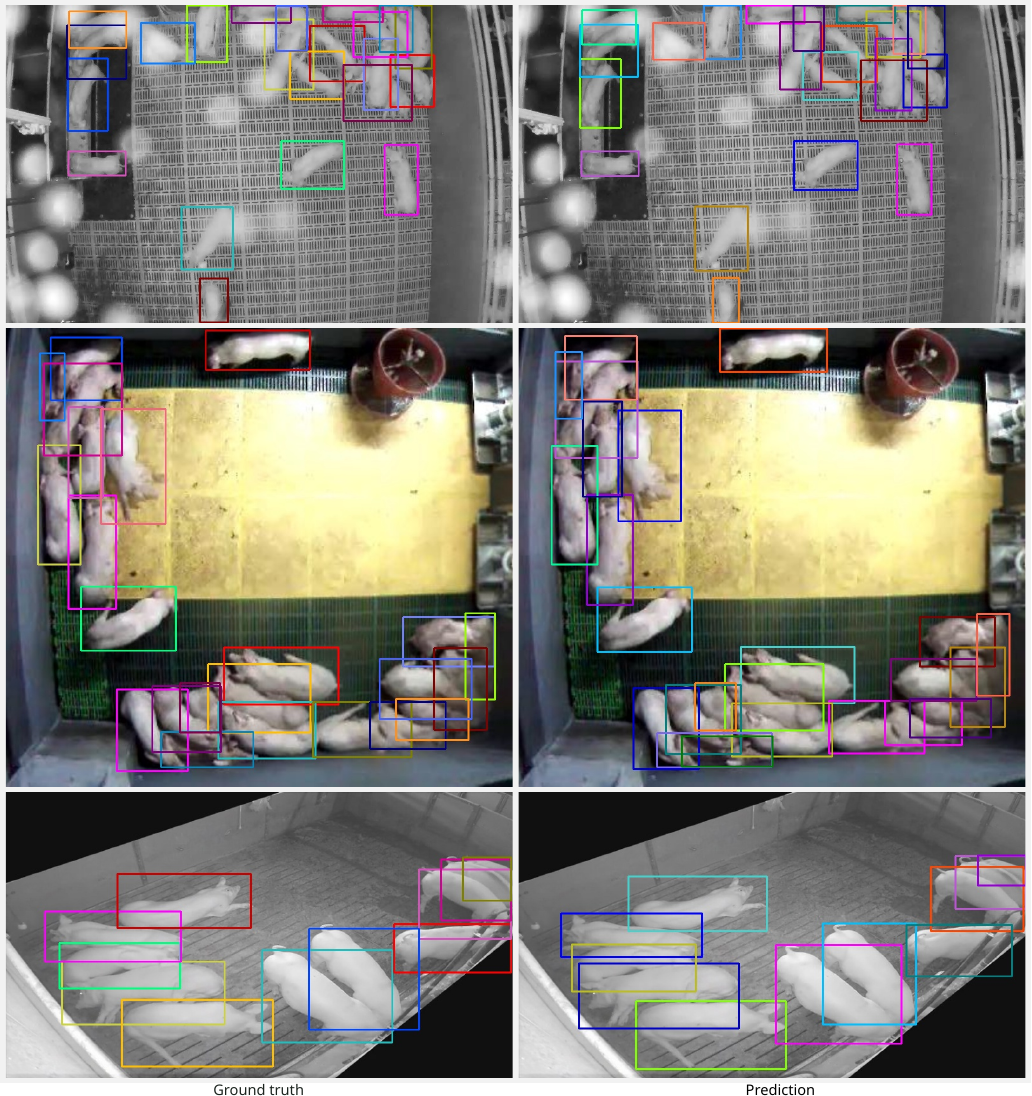}
    \caption{\revchanged{Visualization of ground truth and predicted bounding boxes for three pens that were not part of the training dataset. Top shows an example from the PigDetect test set, middle a frame from the test video published by \citet{yu2025fto}, and bottom an image from the object detection test set published by \citet{melfsen2023describing}. All predicted bounding boxes of the Co-DINO object detector (see \secref{sec:bench_ob_imp}) with a score $\geq0.2$ are visualized. The three images were selected for visualization because they depict challenging scenarios with poor visibility, piling and occlusions. The detection results were not examined prior to selecting the images. For reference, we also provide the names of the respective images in the published datasets: PigDetect (danuma\_1596.jpg), \citet{melfsen2023describing} (931-0035-0037\_3100.jpg), \citet{yu2025fto} (jochiwon5M\_00000865.jpg). The video frame from \citet{yu2025fto} was rotated $90^{\circ}$ clockwise to better fit the page layout.
}}
    \label{fig:app_det_global}
\end{figure*}

\clearpage
\newpage

\section[\appendixname~\thesection]{Tracking datasets}\label{app:sec_track}

\begin{figure*}[h]        % ! = override constraints, h = right here
  \centering
  \includegraphics[width=\textwidth]{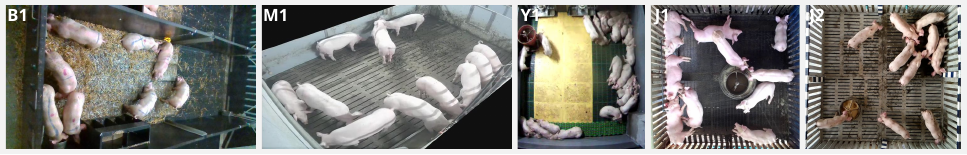}
  \caption{Example images from all unique pen environments contained in the publicly available annotated multi-object tracking datasets presented in \secref{sec:track_char}. The character in the top left corner of each image indicates the initial of the first author of the corresponding work (B\cite{bergamini_extracting_2021}, M\cite{melfsen2023describing}, Y\cite{yu2025fto}, J\cite{jaoukaew2024robust}). The unique pen environments presented in each work are then numbered starting from one.}
  \label{fig:app_track_third}
\end{figure*}

\begin{figure*}[b]        % ! = override constraints, h = right here
  \centering
  \includegraphics[width=\textwidth]{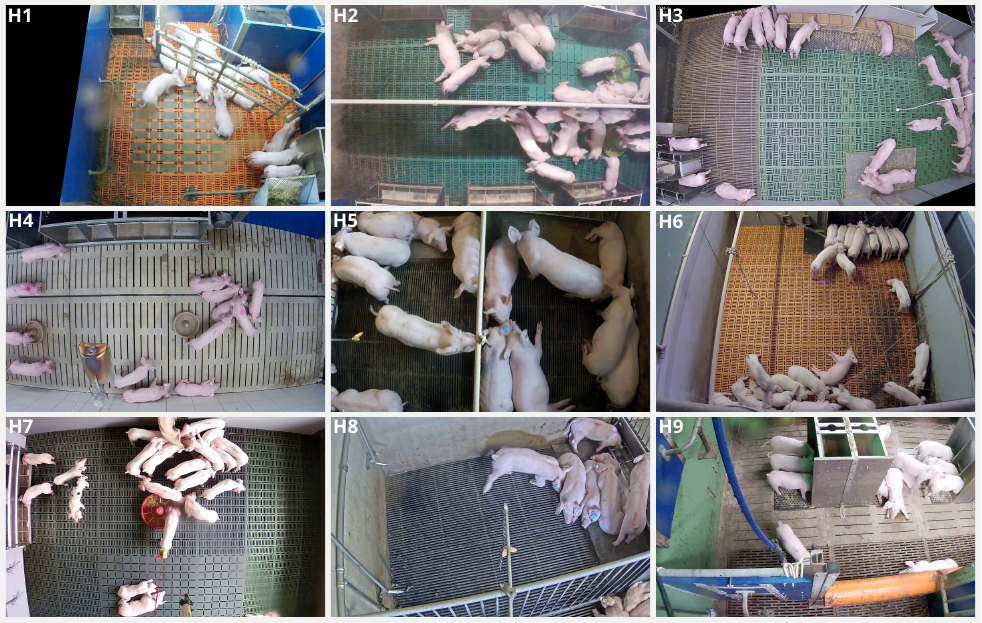}
  \caption{Example images from all unique pen environments contained in the PigTrack dataset. Consistent with other figures in the appendix, the character in the top left corner of each image indicates the initial of the first author of the work in which the dataset was introduced (H for Henrich since the dataset was annotated and released in this work). The unique pen environments are then numbered starting from one. Apart from different pen environments, PigTrack includes videos from different camera angles. These are not depicted here.}
  \label{fig:app_track_ours}
\end{figure*}

See \tab{tab:detailed_pigtrack} and Fig. \ref{fig:app_track_third} and \ref{fig:app_track_ours} on the next two pages.

\newpage

\begin{sidewaystable*}           % full-width, rotated 90°
  \centering
  \caption{Video-level characteristics of the PigTrack dataset. \textit{SeqNum} corresponds to the name of the video in the publicly available dataset. \textit{PenID} corresponds to the pen numbering introduced in \fig{fig:app_track_ours}.}
  \label{tab:detailed_pigtrack}
  \small
  \setlength{\tabcolsep}{4pt}

  % ---- first 40 rows ------------------------------
  \begin{minipage}[t]{0.5\textheight}   % width measured in "page-height"
    \centering
    \begin{tabular}{l c c c c c c c c}
      \toprule
      SeqNum & FPS & \#Frames & Resolution & PenID & Setting & \#Pigs & \#BBoxes & Split \\
      \midrule
        1 & 10 & 149 & 1280x800 & 1 & Day & 13 & 1935 & val \\
        2 & 15 & 985 & 1280x800 & 2 & Day & 26 & 25610 & test \\
        3 & 10 & 246 & 1280x800 & 2 & Night & 26 & 6396 & train \\
        4 & 10 & 340 & 1280x800 & 2 & Night & 26 & 8776 & test \\
        5 & 10 & 230 & 1920x1080 & 3 & Day & 24 & 5505 & test \\
        6 & 10 & 292 & 1920x1080 & 3 & Day & 24 & 7008 & val \\
        7 & 15 & 193 & 1280x800 & 2 & Night & 26 & 5018 & val \\
        8 & 10 & 697 & 1280x800 & 4 & Day & 11 & 7667 & val \\
        9 & 10 & 151 & 1280x800 & 1 & Day & 9 & 1356 & train \\
        10 & 15 & 443 & 1280x800 & 5 & Day & 14 & 6202 & test \\
        11 & 5 & 123 & 1280x800 & 4 & Day & 14 & 1610 & test \\
        12 & 10 & 136 & 1280x800 & 6 & Night & 23 & 3128 & train \\
        13 & 10 & 406 & 1280x800 & 7 & Day & 24 & 9744 & train \\
        14 & 15 & 275 & 1280x800 & 4 & Day & 13 & 3575 & train \\
        15 & 10 & 292 & 1920x1080 & 3 & Day & 24 & 7008 & val \\
        16 & 10 & 431 & 1280x800 & 6 & Night & 23 & 9913 & train \\
        17 & 10 & 620 & 1280x800 & 4 & Night & 14 & 8680 & test \\
        18 & 15 & 450 & 1280x800 & 8 & Night & 7 & 3150 & test \\
        19 & 10 & 97 & 1280x800 & 2 & Night & 26 & 2519 & train \\
        20 & 15 & 451 & 1280x800 & 4 & Day & 13 & 5863 & val \\
        21 & 10 & 641 & 1280x800 & 6 & Night & 23 & 14654 & test \\
        22 & 10 & 599 & 1280x800 & 4 & Night & 12 & 6630 & val \\
        23 & 10 & 560 & 1920x1080 & 3 & Night & 24 & 13427 & train \\
        24 & 15 & 450 & 1280x800 & 8 & Night & 7 & 3106 & train \\
        25 & 15 & 276 & 1280x800 & 5 & Day & 14 & 3864 & train \\
        26 & 10 & 334 & 1280x800 & 2 & Night & 26 & 8684 & train \\
        27 & 10 & 393 & 1280x800 & 7 & Day & 24 & 8941 & train \\
        28 & 10 & 60 & 1280x800 & 4 & Day & 12 & 720 & train \\
        29 & 5 & 161 & 1280x800 & 4 & Day & 14 & 2254 & val \\
        30 & 10 & 273 & 1280x800 & 2 & Day & 26 & 7098 & test \\
        31 & 10 & 291 & 1920x1080 & 3 & Day & 24 & 6984 & val \\
        32 & 10 & 291 & 1920x1080 & 3 & Day & 24 & 6984 & val \\
        33 & 10 & 291 & 1920x1080 & 3 & Day & 21 & 6111 & val \\
        34 & 10 & 1155 & 1280x800 & 7 & Night & 25 & 28477 & val \\
        35 & 10 & 189 & 1280x800 & 9 & Day & 25 & 4157 & train \\
        36 & 15 & 195 & 1280x800 & 2 & Night & 26 & 5040 & train \\
        37 & 15 & 450 & 1280x800 & 8 & Day & 7 & 3150 & val \\
        38 & 10 & 90 & 1920x1080 & 3 & Day & 24 & 2160 & test \\
        39 & 10 & 140 & 1920x1080 & 3 & Day & 23 & 3220 & train \\
        40 & 10 & 296 & 1280x800 & 6 & Day & 23 & 6808 & test \\
      \bottomrule
    \end{tabular}
  \end{minipage}\hfill
  % ---- second 40 rows -----------------------------
  \begin{minipage}[t]{0.5\textheight}
    \centering
    \begin{tabular}{l c c c c c c c c}
      \toprule
      SeqNum & FPS & \#Frames & Resolution & PenID & Setting & \#Pigs & \#BBoxes & Split \\
      \midrule
        41 & 10 & 292 & 1920x1080 & 3 & Day & 24 & 7008 & train \\
        42 & 10 & 530 & 1920x1080 & 3 & Night & 24 & 12720 & test \\
        43 & 10 & 110 & 1920x1080 & 3 & Day & 23 & 2526 & train \\
        44 & 10 & 432 & 1280x800 & 2 & Day & 26 & 11231 & train \\
        45 & 15 & 450 & 1280x800 & 8 & Day & 7 & 3150 & train \\
        46 & 10 & 251 & 1280x800 & 7 & Day & 25 & 6275 & train \\
        47 & 15 & 450 & 1280x800 & 8 & Day & 7 & 3150 & val \\
        48 & 10 & 320 & 1280x800 & 6 & Day & 23 & 7360 & val \\
        49 & 10 & 217 & 1280x800 & 6 & Night & 23 & 4988 & test \\
        50 & 10 & 304 & 1280x800 & 2 & Night & 26 & 7900 & train \\
        51 & 10 & 300 & 1920x1080 & 3 & Day & 24 & 7200 & val \\
        52 & 15 & 195 & 1280x800 & 8 & Night & 7 & 1365 & train \\
        53 & 10 & 330 & 1920x1080 & 3 & Day & 24 & 7920 & train \\
        54 & 10 & 284 & 1280x800 & 6 & Day & 23 & 6475 & test \\
        55 & 10 & 98 & 1280x800 & 9 & Day & 12 & 1176 & test \\
        56 & 10 & 300 & 1920x1080 & 3 & Day & 24 & 7200 & val \\
        57 & 10 & 300 & 1920x1080 & 3 & Day & 24 & 7200 & train \\
        58 & 10 & 605 & 1280x800 & 7 & Night & 24 & 14520 & test \\
        59 & 10 & 80 & 1920x1080 & 3 & Night & 24 & 1920 & train \\
        60 & 10 & 350 & 1280x800 & 9 & Day & 10 & 3499 & test \\
        61 & 10 & 343 & 1280x800 & 2 & Night & 26 & 8918 & train \\
        62 & 10 & 336 & 1280x800 & 7 & Day & 24 & 8064 & val \\
        63 & 10 & 224 & 1280x800 & 1 & Day & 13 & 2912 & train \\
        64 & 10 & 352 & 1280x800 & 4 & Night & 12 & 4224 & train \\
        65 & 10 & 298 & 1280x800 & 2 & Day & 26 & 7748 & test \\
        66 & 10 & 180 & 1280x800 & 1 & Day & 17 & 3060 & val \\
        67 & 10 & 195 & 1280x800 & 7 & Day & 21 & 4079 & test \\
        68 & 10 & 149 & 1280x800 & 1 & Day & 13 & 1937 & test \\
        69 & 10 & 391 & 1280x800 & 2 & Day & 26 & 10166 & val \\
        70 & 10 & 291 & 1280x800 & 6 & Day & 23 & 6693 & train \\
        71 & 10 & 350 & 1280x800 & 4 & Night & 14 & 4884 & test \\
        72 & 10 & 561 & 1280x800 & 7 & Night & 25 & 14025 & test \\
        73 & 15 & 450 & 1280x800 & 8 & Day & 10 & 4500 & train \\
        74 & 15 & 330 & 1280x800 & 8 & Day & 10 & 3300 & test \\
        75 & 10 & 167 & 1280x800 & 9 & Day & 17 & 2794 & train \\
        76 & 15 & 361 & 1280x800 & 4 & Day & 14 & 5004 & train \\
        77 & 10 & 191 & 1280x800 & 7 & Day & 23 & 4363 & test \\
        78 & 10 & 290 & 1920x1080 & 3 & Night & 24 & 6960 & test \\
        79 & 10 & 298 & 1280x800 & 7 & Day & 25 & 7450 & train \\
        80 & 10 & 390 & 1280x800 & 6 & Day & 23 & 8965 & train \\
      \bottomrule
    \end{tabular}
  \end{minipage}
\end{sidewaystable*}

\cleardoublepage

% bibliography
\bibliographystyle{elsarticle-num-names}
\bibliography{literature.bib}

\end{document}